\documentclass[11pt]{article}

\usepackage[utf8]{inputenc}
\usepackage[T1]{fontenc}
\usepackage{amsmath,amssymb,amsfonts}
\usepackage{graphicx}
\usepackage{booktabs}
\usepackage{multirow}
\usepackage{hyperref}
\usepackage{algorithm}
\usepackage{tcolorbox}
\tcbuselibrary{breakable}
\usepackage{algpseudocode}
\usepackage{xcolor}
\usepackage{listings}
\usepackage{enumitem}
\usepackage{natbib}
\usepackage[margin=1in]{geometry}
\usepackage{caption}
\usepackage{tikz}
\usetikzlibrary{shapes,arrows,positioning,fit,calc}

\hypersetup{
    colorlinks=true,
    linkcolor=blue!70!black,
    citecolor=green!50!black,
    urlcolor=blue!60!black
}

\newcommand{\ours}{\textsc{Opti-Agent-Bench}}
\newcommand{\orac}{\textsc{ORAC}}

\title{\textbf{Opti-Agent-Bench: Benchmarking End-to-End Optimization R\&D Agents\\ on Real-World Business Problems}}

\author{
\begin{tabular}{c}
{\large
Yongchang Fu$^{1}$,
Xinjie Huang$^{1,2}$,
Chengjun Dai$^{1}$,
Chengzhe Feng$^{1}$,
Junshao Zhang$^{1}$,
Hong Zhu$^{1}$}\\[0.35em]
{\small $^{1}$Ding Talk, Alibaba Group, Hangzhou, China}\\
{\small $^{2}$Zhejiang University, Hangzhou, China}\\[0.35em]
{\small \texttt{\{fuyongchang.fyc, chengjun.dcj, fengchengzhe.fcz, jiushao.zhangjs, yisu\}@alibaba-inc.com}}\\
{\small \texttt{xinjiehuang@zju.edu.cn}}
\end{tabular}
}

\date{}

\begin{document}

\maketitle

\begin{abstract}
LLM-based agents are increasingly deployed to solve optimization problems, yet existing benchmarks evaluate them on pre-structured mathematical formulations that bypass the most critical challenge: \emph{translating complex business requirements into correct models and solve efficiently}. We introduce \ours{}, an end-to-end benchmark that evaluates Large Language Models (LLMs) across the complete optimization R\&D pipeline, from understanding business-language descriptions through mathematical modeling, algorithm selection, and code implementation, to solution report generation. Our design rests on three pillars: (1) \textbf{business-semantic authenticity} with anti-template traps that defeat pattern matching; (2) \textbf{modular evaluation with cross-module consistency checking} across Problem Understanding, Formal Modeling, Implementation, and Reporting; and (3) the \orac{} \textbf{bi-level validity framework} that simultaneously ensures task quality and scoring integrity. Across several industrial-scale tasks spanning integer programming, robust optimization, stochastic programming, and non-convex optimization, we expose critical failure modes of current models, including constraint omission, model-code inconsistency, and report-implementation divergence, that remain invisible under conventional single-metric evaluation.
\end{abstract}

\section{Introduction}
\label{sec:intro}

\begin{figure}
    \centering
    \includegraphics[width=0.7\linewidth]{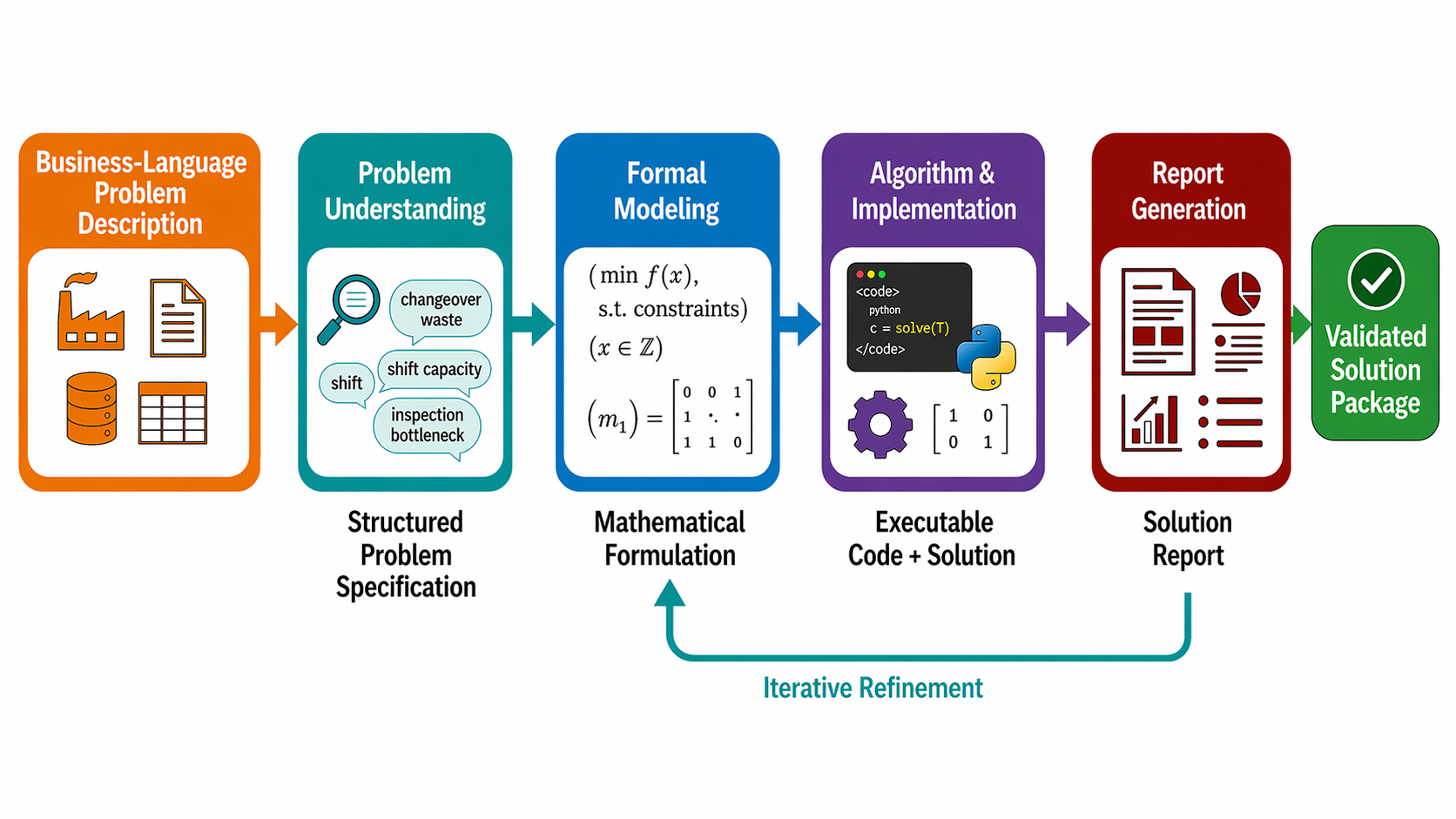}
    \caption{A Pipeline for Solving Optimization Problems in Real-World Business Scenarios.}
    \label{fig:RDppl}
\end{figure}

Operations research and mathematical optimization underpin decision-making across industries---from supply chain logistics and production scheduling to financial portfolio design and energy network management. In practice, however, the path from a business stakeholder's requirements to a deployed optimization solution is far from straightforward. It demands a complete R\&D pipeline: understanding the business context, identifying the underlying optimization structure, formulating a rigorous mathematical model, selecting and implementing an appropriate solution algorithm, and communicating the results through a clear, faithful report (see Figure~\ref{fig:RDppl}). Each link in this chain requires deep domain knowledge, mathematical sophistication, and engineering discipline.

The rapid advancement of LLM-based agents has generated considerable excitement about automating this pipeline. Systems such as OptiMUS~\citep{optimus2023} have demonstrated that LLM agents can, in favorable conditions, translate natural-language problem descriptions into solver code and iteratively debug their solutions. Meanwhile, foundation model approaches for mixed-integer linear programming (MILP)~\citep{milpevolve2025} and knowledge-augmented formulation frameworks~\citep{pengautomatic2025} suggest that integrating LLMs with optimization tools can yield practically useful results. Yet a fundamental question remains largely unanswered: \emph{How reliably can these agents handle the full complexity of real-world optimization R\&D?}

\subsection{The Evaluation Gap}

Current benchmarks for optimization agents suffer from three interrelated limitations that systematically overestimate agent capabilities.

\paragraph{Over-Structured Problem Descriptions.}
Most existing benchmarks---including NLP4LP~\citep{optimus2023}, OptiBench~\citep{optibench2025}, and the NL4Opt competition~\citep{nl4opt2023}---present problems that are already substantially structured. Task descriptions often use academic terminology (``minimize the total weighted tardiness,'' ``subject to capacity constraints''), making the optimization type immediately recognizable. Agents can exploit this by pattern-matching to known templates (TSP, knapsack, facility location) rather than genuinely understanding the business requirements. As MIPLIB-NL~\citep{miplbnl2026} demonstrated, fine-tuned models that achieve 80--95\% accuracy on such benchmarks collapse to near-zero performance when confronted with industrial-scale problems whose structure must be inferred from realistic descriptions.

\paragraph{Single-Metric Evaluation.}
Conventional benchmarks evaluate agents primarily by the objective function value of their solutions or binary pass/fail against solver-verified answers~\citep{optmath2025,optibench2025}. This single-metric approach masks a host of critical errors: an agent may arrive at a reasonable objective value through an incorrect model that happens to produce a feasible solution by coincidence, or through reward hacking that exploits evaluation loopholes. ORQA~\citep{orqa2025} showed that even when LLMs appear to understand optimization problems (77.2\% accuracy on multiple-choice questions), they fail at the deeper reasoning required for actual formulation---and that chain-of-thought prompting, counterintuitively, can \emph{degrade} performance by triggering hallucinations.

\paragraph{Pipeline Fragmentation.}
No existing benchmark evaluates the complete optimization R\&D pipeline facing real-world business problems. ORQA~\citep{orqa2025} tests conceptual understanding but not formulation. MIPLIB-NL~\citep{miplbnl2026} and OptMATH~\citep{optmath2025} test formulation but not iterative refinement or reporting. ALE-Bench~\citep{alebench2025} evaluates algorithmic engineering with iterative improvement but does not require natural-language-to-model translation. OptiMUS evaluates an end-to-end agent but only on 52 textbook-scale problems. The result is that no benchmark captures the full spectrum of capabilities---and especially the \emph{cross-stage consistency}---required of a genuine optimization R\&D pipeline.

\subsection{Our Approach}

\begin{figure}
    \centering
    \includegraphics[width=0.5\linewidth]{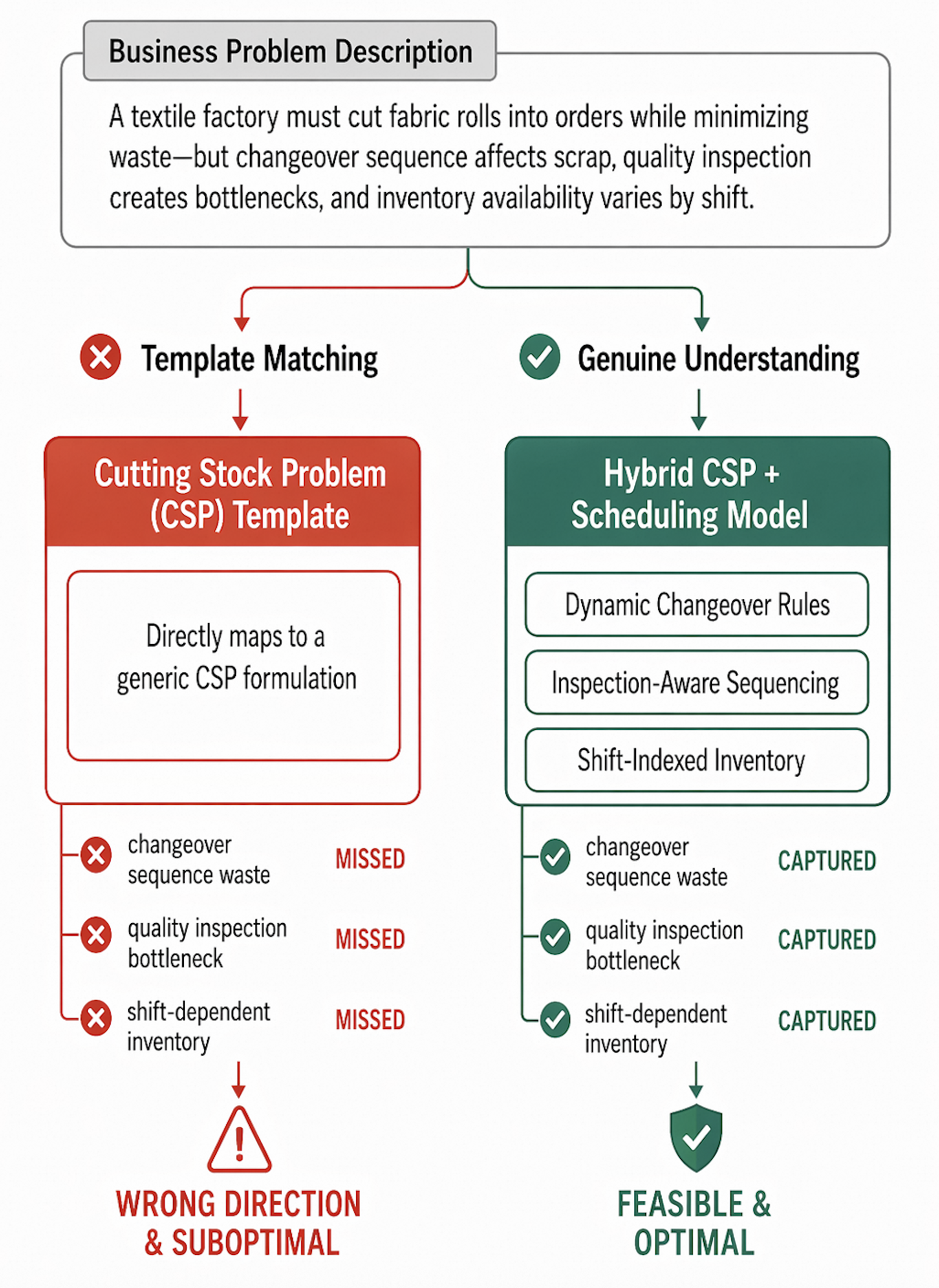}
    \caption{Anti-Template Trap: Conceptual Comparison of Agent Reasoning Paths}
    \label{fig:antitemp}
\end{figure}

\ours{} addresses these gaps through three complementary innovations:

\textbf{(1) Business-grounded task design.} Every task in our benchmark is described exclusively in business language---the language of production managers, logistics planners, and financial analysts---without academic optimization jargon. Tasks incorporate anti-template traps: features that cause standard model templates to produce infeasible or suboptimal solutions (see Figure~\ref{fig:antitemp}). Hidden constraints, coupled decision variables, and domain-specific terminology force agents to perform genuine business understanding and creative modeling, not mere pattern recognition.

\textbf{(2) Modular pipeline evaluation with consistency checking.} We decompose the optimization R\&D pipeline into four evaluable modules: \emph{Problem Understanding}, \emph{Formal Modeling}, \emph{Algorithm Design \& Implementation}, and \emph{Report Generation}. Each module is independently scored, enabling fine-grained diagnosis of agent capabilities. Crucially, we introduce \emph{cross-module consistency metrics} that verify whether an agent's mathematical model faithfully reflects its stated problem understanding, whether its code faithfully implements its model, and whether its report accurately describes its implementation. This modular architecture also supports partial evaluation: researchers can assess any contiguous subset of modules along with their inter-module consistency.

\textbf{(3) Bi-level validity framework (\orac{}).} The \orac{} checklist ensures evaluation rigor from two directions simultaneously. \emph{Problem Formulation Validity} verifies that tasks are genuinely challenging---resistant to template matching, inclusive of hidden constraints, and grounded in authentic business semantics. \emph{Solution Validity} verifies that scoring metrics truly reflect solution quality---distinguishing feasibility from correctness, objective values from model fidelity, and code runnability from implementation integrity.

\subsection{Contributions}

The contributions of this work are as follows:

\begin{enumerate}[leftmargin=*]
    \item We propose \ours{}, the first benchmark that evaluates optimization R\&D agents on the \emph{complete} pipeline from business-language requirements to solution reports, covering nine industrial-scale optimization scenarios across diverse problem types (integer programming, robust optimization, stochastic programming, Markov decision processes, second-order cone programming, and non-convex optimization).

    \item We introduce a \textbf{modular evaluation architecture} that independently assesses Problem Understanding, Formal Modeling, Implementation, and Reporting, together with novel \emph{cross-module consistency metrics} that detect misalignment between adjacent pipeline stages. This architecture supports both full-pipeline and partial-module evaluation.

    \item We develop \orac{}, a \textbf{bi-level validity framework} comprising a Problem Formulation Validity checklist for task design and a Solution Validity checklist for scoring integrity, ensuring that both the benchmark tasks and the evaluation metrics meet rigorous scientific standards.

    \item Through systematic experiments, we \textbf{identify and categorize critical failure modes} of current LLM agents---including constraint omission under business language, model--code inconsistency, scalability collapse, and report--implementation divergence---that are invisible under conventional single-metric evaluation.
\end{enumerate}

\section{Literature Review}
\label{sec:related}

We organize related work into three streams: general-purpose agent benchmarks that establish evaluation paradigms, optimization-specific benchmarks and datasets, and LLM-based optimization systems.

\subsection{General-Purpose Agent Benchmarks}

The past two years have witnessed a proliferation of benchmarks for evaluating LLM-based agents across diverse domains. SWE-Bench~\citep{swebench2024} evaluates code agents on real-world GitHub issue resolution, requiring models to navigate large codebases and produce correct patches verified by unit tests. Its key insight---that real-world tasks are dramatically harder than synthetic ones---has profoundly influenced benchmark design. ALE-Bench~\citep{alebench2025} focuses on algorithmic engineering in competitive programming, introducing score-based (rather than binary) evaluation and demonstrating the value of iterative refinement over long time horizons. CL-Bench~\citep{clbench2026} evaluates context learning---the ability to acquire and apply new knowledge from complex in-context materials---revealing that even frontier models solve fewer than 25\% of context-dependent tasks. RE-Bench~\citep{rebench2024} and MLE-Bench~\citep{mlebench2024} assess research engineering and machine learning engineering capabilities, respectively. In the code generation space, BigCodeBench~\citep{bigcodebench2025} introduces 1,140 tasks covering 139 libraries to evaluate compositional code generation with diverse function calls, while LiveCodeBench~\citep{livecodebench2025} proposes a continuously updated benchmark using new competitive programming problems to prevent data contamination. DSBench~\citep{dsbench2025} evaluates end-to-end data science agents on 466 data analysis and 74 modeling tasks from Kaggle competitions, offering a holistic assessment of multi-step analytical pipelines.

These benchmarks collectively establish important evaluation principles: execution-based verification over surface-level output matching, real-world grounding over synthetic simplification, and multi-faceted assessment over single-metric scoring. Beyond formal benchmarks, Knuth~\citep{knuth2026} documents a striking case study in which an LLM was guided through over 30 explorations to solve an open combinatorial problem on directed Hamiltonian cycle decomposition, demonstrating that while LLMs can discover non-trivial mathematical constructions, the process required extensive human coaching and the model eventually degraded, underscoring the gap between assisted exploration and autonomous R\&D. In the multimodal reasoning space, MATH-V~\citep{mathv2024} reveals that even frontier models achieve only 22.76\% accuracy on competition-level mathematical problems, with reasoning errors as the dominant failure mode. FrontierMath~\citep{frontiermath2024} pushes the boundary further with research-level mathematics problems created by 60+ professional mathematicians, where frontier models solve fewer than 2\% of problems. Omni-MATH~\citep{omnimath2025} provides a universal olympiad-level mathematical benchmark with broad topic coverage and fine-grained difficulty ratings, confirming that mathematical reasoning remains a significant bottleneck for current LLMs. Recent work on test-time compute scaling~\citep{snell2025} demonstrates that optimally allocating additional inference-time computation can match or exceed the benefits of larger model parameters, suggesting promising directions for improving LLM reasoning on complex optimization tasks. \ours{} inherits the evaluation principles established by these works while specializing them for the unique challenges of optimization R\&D.

\subsection{Optimization-Specific Benchmarks and Datasets}

Benchmarks specifically targeting LLM capabilities in optimization span a spectrum from conceptual understanding to code generation.

At the understanding end, ORQA~\citep{orqa2025} provides 1,513 multiple-choice questions testing whether LLMs can identify optimization model components (sets, parameters, variables, objectives, constraints) from natural-language descriptions. While valuable for assessing conceptual grasp, ORQA does not require agents to generate formulations, write code, or produce solutions.

At the formulation end, several benchmarks evaluate the translation from natural language to mathematical models. The NLP4LP dataset~\citep{optimus2023} contains 52 LP/MILP problems from textbooks with expert-verified formulations. Li et al.~\citep{limipsynth2023} propose a three-phase framework integrating LLMs with constraint classification and constraint templates to synthesize MILP models from natural language, achieving 86.67\% problem accuracy on an extended NL4Opt dataset. OptiBench~\citep{optibench2025} expands to 605 problems including nonlinear programming and tabular data, introducing the ReSocratic synthesis method for generating training data. OptMATH~\citep{optmath2025} contributes a scalable bidirectional data synthesis framework, producing approximately 200K training triplets with 99.6\% equivalence accuracy verified through solver-based rejection sampling. OPT-Engine~\citep{optengine2026} introduces an extensible benchmark with controllable complexity scaling across 10 canonical LP/MIP tasks, revealing that tool-integrated reasoning with external solvers is more robust than pure-text reasoning as complexity increases and that constraint formulation is the primary performance bottleneck. MIPLIB-NL~\citep{miplbnl2026} takes a fundamentally different approach: starting from real industrial MILP models in the MIPLIB~2017 repository, experts perform structure-aware reverse construction to generate natural-language descriptions for 223 instances with $10^3$--$10^6$ variables. This work delivers the sobering finding that models achieving high accuracy on toy benchmarks fail catastrophically at industrial scale (0--5.5\% for fine-tuned models). CO-Bench~\citep{cobench2025} evaluates LLM agents specifically on designing algorithms for 36 combinatorial optimization problems across 8 categories, measuring both solution quality and computational efficiency.

Classical optimization instance libraries---TSPLIB, MIPLIB~\citep{miplib2017}, and ORLIB---provide standardized problem instances but present them in mathematical or data-file formats, bypassing the natural-language understanding challenge entirely.

\subsection{LLM-Based Optimization Systems}

Beyond benchmarking, several systems demonstrate LLM agents in optimization roles. OptiMUS~\citep{optimus2023} implements a multi-component pipeline (formulation, code generation, debugging, auto-testing, problem rephrasing) that nearly doubles success rates over naive prompting, though it is evaluated only on 52 textbook problems. Astorga et al.~\citep{astorga2025} formalize the autoformulation problem---the automated translation of natural-language descriptions into formal mathematical programs---and introduce symbolic representation with active acquisition techniques at ICML, establishing a theoretical foundation for the field. LLMOPT~\citep{llmopt2025} proposes a unified learning-based framework that teaches LLMs to both define and solve general optimization problems, improving generalization across diverse problem types. OptiTree~\citep{optitree2025} introduces a tree-search-based hierarchical thought generation approach that significantly improves LLM performance on multi-step optimization modeling tasks. OR-LLM-Agent~\citep{orllmagent2025} presents a multi-agent system for end-to-end operations research that separates modeling, coding, and debugging roles across specialized LLM agents with reasoning capabilities. ORLM~\citep{orlm2025} proposes a semi-automated data synthesis and fine-tuning framework for training open-source LLMs to perform optimization modeling, published in the flagship INFORMS \emph{Operations Research} journal. Talebi~\citep{talebi2025} extends LLM-based formulation to stochastic optimization problems, introducing a soft-scoring metric for partial correctness. Peng et al.~\citep{pengautomatic2025} demonstrate knowledge-augmented MILP construction for multi-robot scheduling using locally deployed models, achieving 82\% modeling accuracy on aircraft manufacturing constraints.

In the domain of LLM-driven algorithm and heuristic design, Romera-Paredes et al.~\citep{funsearch2024} demonstrate in \emph{Nature} that LLM-guided evolutionary program search (FunSearch) can make genuine new mathematical discoveries on the cap set problem, establishing a paradigm for using LLMs as scientific discovery tools. Ye et al.~\citep{reevo2024} introduce Reflective Evolution (ReEvo), where LLMs iteratively generate and refine heuristic algorithms using long-term reflection for combinatorial optimization. Cetinkaya et al.~\citep{cetinkaya2025} leverage LLMs to discover novel scheduling heuristics through an evolutionary pipeline, producing interpretable algorithms that compete with exact methods. The LOOP framework~\citep{loop2026} combines neural and symbolic components for planning, achieving 85.8\% success rates through iterative neural-symbolic conversation.

Two comprehensive surveys provide valuable context for this rapidly evolving field. Xiao et al.~\citep{xiaosurvey2025} present a systematic survey at IJCAI categorizing progress at the intersection of optimization modeling and LLMs, covering formulation, solving, and evaluation with a reconstructed leaderboard. Liu et al.~\citep{liusurveyalgo2025} introduce a taxonomy in \emph{ACM Computing Surveys} categorizing how LLMs serve as searchers, generators, and evaluators in algorithm design, spanning heuristic design, meta-heuristics, and combinatorial optimization.

These systems and surveys showcase the rapid growth of LLM agents for optimization but reveal that each system is evaluated in narrow, controlled settings. On the solver acceleration side, Li et al.~\citep{milpreduction2025} propose preference-based model reduction learning that achieves orders-of-magnitude speedups on real-world MILPs, illustrating that ML-based techniques can complement traditional solvers but still require correctly formulated models as input. None of these systems is assessed on the full business-language-to-report pipeline that characterizes real-world optimization R\&D.

\subsection{Positioning of \ours{}}

Table~\ref{tab:positioning} summarizes the landscape. Each existing benchmark captures at most one or two dimensions of the optimization R\&D pipeline. The closest prior work is MIPLIB-NL~\citep{miplbnl2026}, which shares our commitment to business-language descriptions and industrial scale. However, MIPLIB-NL evaluates only the formulation-to-execution path (whether generated code produces the correct solver output) without assessing the agent's problem understanding, its modeling rationale, the consistency between its stated model and its code, or its ability to communicate results through a report. \ours{} is the first to integrate all dimensions: business-language input, mathematical modeling, code implementation, solution validation, report generation, and cross-stage consistency verification.

\begin{table}[t]
\centering
\caption{Positioning of \ours{} relative to existing benchmarks. \textbf{BL}: business-language input (\checkmark\ = full, $\circ$ = partial/semi-structured); \textbf{FM}: formal modeling evaluation; \textbf{CE}: code/execution evaluation; \textbf{RE}: report evaluation; \textbf{CC}: cross-module consistency; \textbf{AT}: anti-template design; \textbf{IS}: industrial scale.}
\label{tab:positioning}
\small
\begin{tabular}{lccccccc}
\toprule
\textbf{Benchmark} & \textbf{BL} & \textbf{FM} & \textbf{CE} & \textbf{RE} & \textbf{CC} & \textbf{AT} & \textbf{IS} \\
\midrule
ORQA~\citep{orqa2025}         & \checkmark & -- & -- & -- & -- & -- & -- \\
NLP4LP~\citep{optimus2023}    & $\circ$ & \checkmark & \checkmark & -- & -- & -- & -- \\
OptiBench~\citep{optibench2025} & $\circ$ & \checkmark & \checkmark & -- & -- & -- & -- \\
OptMATH~\citep{optmath2025}   & $\circ$ & \checkmark & \checkmark & -- & -- & -- & -- \\
OPT-Engine~\citep{optengine2026} & $\circ$ & \checkmark & \checkmark & -- & -- & -- & -- \\
CO-Bench~\citep{cobench2025}  & -- & -- & \checkmark & -- & -- & -- & -- \\
MIPLIB-NL~\citep{miplbnl2026} & \checkmark & \checkmark & \checkmark & -- & -- & -- & \checkmark \\
ALE-Bench~\citep{alebench2025}  & -- & -- & \checkmark & -- & -- & -- & -- \\
SWE-Bench~\citep{swebench2024} & -- & -- & \checkmark & -- & -- & -- & \checkmark \\
\midrule
\ours{} (Ours)               & \checkmark & \checkmark & \checkmark & \checkmark & \checkmark & \checkmark & \checkmark \\
\bottomrule
\end{tabular}
\end{table}

\section{From Structured Problems to Business Reality}
\label{sec:motivation}

The central innovation of \ours{} lies in recognizing that the most critical---and most neglected---challenge in optimization R\&D is not solving a well-formulated mathematical program, but \emph{arriving at the correct formulation from a messy, ambiguous, domain-specific business description}. This section articulates why this distinction matters and how it shapes our benchmark design.

\subsection{The Formulation Gap}

Consider the contrast between a typical benchmark problem and a real business requirement:

\begin{quote}
\textbf{Typical benchmark:} ``Minimize the total transportation cost for shipping goods from $m$ warehouses to $n$ customers, subject to supply and demand constraints.''

\textbf{Real business requirement:} ``Our custom roll-material processing line runs three shifts. Each shift can process up to 500 meters of packaging-film material, but changeover between material grades takes 45 minutes and wastes approximately 12 meters of raw material. We need to fulfill next week's custom cutting orders while minimizing waste. Oh, and the quality inspection team can only process 200 meters per shift---anything beyond that gets queued to the next shift.''
\end{quote}

The first description is essentially a mathematical formulation in natural language: the optimization type (LP/transportation), decision variables, objective, and constraints are all explicit. An agent can solve it through template matching without genuine understanding. The second requires the agent to: (1) recognize this as a cutting-stock/production-scheduling hybrid; (2) identify that changeover time and material waste create coupled constraints; (3) realize that the quality inspection bottleneck introduces a secondary capacity constraint that interacts non-trivially with the production schedule; and (4) determine that the time-dependent nature of shifts makes this fundamentally different from a static optimization problem.

This \emph{formulation gap}---the cognitive distance between business language and mathematical structure---is where real optimization expertise resides, and where current agents most critically fail.

\subsection{Why Template Matching Fails on Real Problems}

MIPLIB-NL~\citep{miplbnl2026} provides compelling evidence that template matching is the dominant strategy of current LLM-based optimization systems, and that it fails catastrophically on non-template problems. Fine-tuned models (SIRL-32B, OptMATH-7B) that achieve 80--95\% accuracy on existing benchmarks score 0--5.5\% on MIPLIB-NL's industrial instances. The failure mode analysis reveals that 48.3\% of errors are \emph{modeling-level}---missing constraints, incomplete index sets, incorrect loop structures---precisely the errors that would arise from attempting to fit a template to a problem it does not match.

Our benchmark deliberately amplifies this challenge through \emph{anti-template design}: each task contains features that cause standard model templates to produce infeasible or suboptimal solutions. For instance, a vehicle routing task might include time-dependent travel speeds that invalidate the standard distance-matrix assumption, or a portfolio optimization task might incorporate regulatory constraints that transform a convex problem into a non-convex one. These traps are invisible to pattern-matching agents but obvious to a human expert who carefully reads the business requirements.

\subsection{The End-to-End Optimization R\&D Pipeline}

Real-world optimization R\&D is not a single translation step from natural language to mathematical model. It is a multi-stage pipeline where each stage builds on, and must be consistent with, the preceding one:

\begin{enumerate}[leftmargin=*]
    \item \textbf{Problem Understanding}: Identifying the business objective, decision variables, constraints (explicit and implicit), and the appropriate problem class from a natural-language description.
    \item \textbf{Formal Modeling}: Translating the understood problem into a rigorous mathematical formulation with precisely defined sets, parameters, variables, objective function, and constraints.
    \item \textbf{Algorithm Design \& Implementation}: Selecting an appropriate solution method (exact solver, heuristic, metaheuristic, decomposition) and implementing it in executable code that faithfully realizes the mathematical model.
    \item \textbf{Report Generation}: Producing a clear, accurate report that explains the modeling decisions, presents the solution, and communicates insights to business stakeholders.
\end{enumerate}

Critically, the quality of the final output depends not only on the quality of each individual stage but also on the \emph{consistency between adjacent stages}. A mathematically correct model is worthless if the code implements a different model. A correct implementation is misleading if the report describes different constraints. This cross-stage consistency is invisible to benchmarks that evaluate only the final solver output.

\section{The ORAC Validity Framework}
\label{chap:orac}

To ensure that \texttt{Opti-Agent-Bench} provides a scientifically rigorous and faithful evaluation of optimization R\&D agents, we develop the \textbf{O}ptimization \textbf{R}\&D \textbf{A}gent \textbf{C}hecklist (ORAC). ORAC is a bi-level validity framework designed to address two complementary questions: \textit{Are our tasks genuinely challenging and representative of real-world complexity?} (Problem Formulation Validity), and \textit{Do our evaluation metrics accurately and robustly measure what truly matters?} (Solution Validity). This framework moves beyond simplistic metrics by establishing a robust foundation for both task design and the assessment process itself.

\subsection{Problem Formulation Validity: Crafting High-Fidelity Tasks}
\label{subsec:pf_validity}

Problem Formulation Validity ensures that each benchmark task genuinely tests an agent's ability to reason, model, and solve complex business problems, rather than merely pattern-matching to known academic templates. This is a direct response to the observation that models performing well on textbook problems often fail catastrophically on industrial-scale tasks \cite{miplbnl2026}. We operationalize this through four interconnected design principles:

\begin{itemize}
    \item[\textbf{TD.1}] \textbf{Business-Semantic Authenticity.} Task descriptions must use authentic business language---the vocabulary of domain practitioners, not optimization textbooks. Phrases like ``changeover downtime,'' ``yield rate,'' and ``regulatory capital buffer'' replace their academic counterparts like ``setup cost,'' ``efficiency parameter,'' and ``budget constraint.'' This forces agents to perform genuine semantic grounding by mapping business concepts to mathematical constructs, a critical step in real-world problem-solving.

    \item[\textbf{TD.2}] \textbf{Anti-Template Design.} Each task must contain at least one structural feature or ``trap'' that causes the most natural template model (e.g., a standard Cutting Stock or Mean-Variance model) to produce a suboptimal or infeasible solution. These traps include non-standard objective functions (e.g., minimizing maximum deviation), hidden coupling between apparently independent constraints, and problem structures that are deliberately misleading (e.g., a problem that looks like vehicle routing but is actually a scheduling problem with routing aspects).

    \item[\textbf{TD.3}] \textbf{Multi-Constraint Coupling.} Tasks must include constraints that interact non-trivially. Satisfying one constraint should affect the feasibility or optimality of others, requiring agents to reason about the system as a coherent whole rather than a mere list of isolated conditions. This tests the agent's ability to build integrated models.

    \item[\textbf{TD.4}] \textbf{Anti-Pattern-Matching.} Task descriptions must avoid keywords that directly reveal the problem type. Instead of stating, ``This is a traveling salesman problem,'' the description presents the underlying business scenario (e.g., ``A field technician must visit all client sites before 5 PM''). This prevents agents from bypassing the crucial formulation stage through simple keyword-triggered retrieval.
\end{itemize}

\subsection{Solution Validity: Principles for Rigorous Evaluation}
\label{sec:solution_validity}

While Problem Formulation Validity ensures the quality of our tasks, Solution Validity ensures the integrity and depth of our scoring. Conventional benchmarks often fall short by equating runnable code with a correct solution or a good objective value with a good model. Solution Validity establishes the conceptual principles that our evaluation system must enforce; the concrete metrics and rubric mechanics that operationalize these principles are detailed in Section~\ref{subsec:metrics}.

\subsubsection{Critical Distinctions}
\label{subsec:critical_distinctions}

The core insight behind Solution Validity is that optimization benchmarks must distinguish between several forms of superficial success that mask genuine capability gaps:

\begin{itemize}
    \item \textbf{Feasible $\neq$ Correct.} A solution may satisfy all constraints of an agent's \textit{wrong} model while violating the constraints of the \textit{correct} model. Evaluation must verify feasibility against the ground-truth Oracle model's constraints, not merely the agent's stated constraints.

    \item \textbf{Good Objective $\neq$ Good Model.} An agent may achieve a competitive objective value through a flawed model---for example, by omitting a critical constraint that happens not to be binding on a specific test instance. Evaluation must inspect the model for structural integrity, detecting such ``coincidental correctness'' that multi-instance testing also helps expose.

    \item \textbf{Runnable Code $\neq$ Correct Implementation.} Code that executes without errors may still implement a different model than the one the agent formally described. Evaluation must perform a ``semantic diff'' between the mathematical formulation and its code implementation, verifying that all stated constraints, objectives, and variable definitions are faithfully translated.

    \item \textbf{Coherent Report $\neq$ Faithful Report.} A report may be well-written yet describe a different optimization approach than what was actually implemented---a form of technical hallucination. Evaluation must cross-reference reported claims against actual execution logs and code structure.
\end{itemize}

\subsubsection{The Evaluator Agent}
\label{subsec:evaluator_agent}

A cornerstone of our framework's scalability and objectivity is the use of an \textbf{Automated Evaluator Agent}. Instead of relying on time-consuming, costly, and potentially inconsistent human expert reviews, we leverage a powerful, specially prompted Large Language Model (Claude Opus 4.6) to act as the evaluator. This Evaluator Agent is given the role of a senior Operations Research expert and is tasked with systematically applying structured rubrics to the complete set of outputs generated by the agent being tested (i.e., its problem analysis, mathematical model, code, and final report). The rubrics provide the Evaluator Agent with concrete, binary-verifiable checkpoints and explicit pass/fail standards, ensuring consistent and reproducible scoring across all tasks and models. The detailed rubric structure and scoring mechanics are described in Section~\ref{subsec:metrics}.

\subsubsection{Design Philosophy: Anti-Template Alignment}
\label{subsec:at_alignment_philosophy}

A distinctive feature of our evaluation design is its explicit coupling with the anti-template traps embedded in each task (TD.2). For every structural feature designed to cause standard templates to fail, the evaluation system contains at least one critical checkpoint that directly tests whether the LLM has fallen into the trap. This design philosophy ensures that an LLM anchoring to a familiar template receives a low score that accurately reflects its failure to engage with the problem's true structure---even if its output appears superficially competent. The mechanism by which this philosophy is implemented through severity-weighted rubric checkpoints is detailed in Section~\ref{subsec:metrics}.

\subsubsection{Illustrative Example: From Principles to Practice}

To demonstrate how the ORAC framework is applied in practice, we provide several comprehensive examples in the appendix. These examples include the full business problem description, the detailed rubrics used for automated evaluation, the pseudo-code for the expert-verified reference model, and a runnable Python prototype. These case studies vividly illustrate our design principles: for instance, the primary anti-template trap (TD.2) in one task requires the agent to move beyond a standard template approach and instead tackle a complex combinatorial optimization problem arising from non-obvious structural features. Correspondingly, a critical-severity checkpoint in the rubric directly enforces this by assessing whether the agent correctly models this structural dependency, thus operationalizing our rigorous standards for Solution Validity.

\section{Modular Evaluation Architecture}
\label{sec:modules}

A distinguishing feature of \ours{} is its modular evaluation architecture, which decomposes the optimization R\&D pipeline into independently assessable modules while maintaining rigorous cross-module consistency checks. This design serves two purposes: it enables fine-grained diagnosis of agent capabilities, and it supports flexible evaluation---researchers can assess any contiguous subset of modules along with their inter-module consistency.

\subsection{Module Decomposition}

The evaluation pipeline comprises four modules, each with clearly defined inputs, outputs, and assessment criteria.

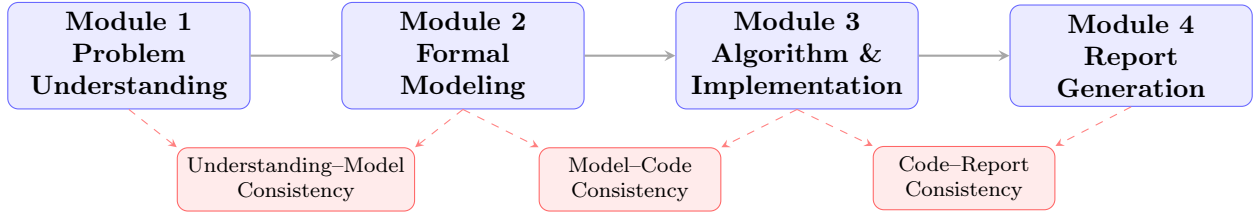
\begin{figure}[h]
\centering
\begin{tikzpicture}[
    node distance=0.3cm and 1.2cm,
    module/.style={rectangle, draw=blue!60, fill=blue!8, rounded corners, minimum width=3.2cm, minimum height=1.2cm, align=center, font=\small\bfseries},
    consistency/.style={rectangle, draw=red!60, fill=red!8, rounded corners, minimum width=2.4cm, minimum height=0.7cm, align=center, font=\scriptsize},
    arrow/.style={->, >=stealth, thick, draw=gray!70}
]
    \node[module] (M1) {Module 1\\Problem\\Understanding};
    \node[module, right=of M1] (M2) {Module 2\\Formal\\Modeling};
    \node[module, right=of M2] (M3) {Module 3\\Algorithm \&\\Implementation};
    \node[module, right=of M3] (M4) {Module 4\\Report\\Generation};

    \draw[arrow] (M1) -- (M2);
    \draw[arrow] (M2) -- (M3);
    \draw[arrow] (M3) -- (M4);

    \node[consistency, below=0.5cm of $(M1.south)!0.5!(M2.south)$] (C12) {Understanding--Model\\Consistency};
    \node[consistency, below=0.5cm of $(M2.south)!0.5!(M3.south)$] (C23) {Model--Code\\Consistency};
    \node[consistency, below=0.5cm of $(M3.south)!0.5!(M4.south)$] (C34) {Code--Report\\Consistency};

    \draw[->, >=stealth, dashed, draw=red!50] (M1.south) -- (C12.north west);
    \draw[->, >=stealth, dashed, draw=red!50] (M2.south) -- (C12.north east);
    \draw[->, >=stealth, dashed, draw=red!50] (M2.south) -- (C23.north west);
    \draw[->, >=stealth, dashed, draw=red!50] (M3.south) -- (C23.north east);
    \draw[->, >=stealth, dashed, draw=red!50] (M3.south) -- (C34.north west);
    \draw[->, >=stealth, dashed, draw=red!50] (M4.south) -- (C34.north east);
\end{tikzpicture}
\caption{Modular evaluation architecture of \ours{}. Blue boxes represent independently assessable modules; red boxes represent cross-module consistency checks between adjacent modules.}
\label{fig:architecture}
\end{figure}

\subsubsection{Module 1: Problem Understanding}
\label{subsec:mod1}

\textit{Input}: Natural-language business problem description with data files.\\
\textit{Output}: Structured problem specification identifying the optimization objective, decision variables, constraint categories, problem class, and any special structural properties.\\
\textit{Assessment}: Evaluated against expert-defined ground truth on completeness (all key elements identified), correctness (elements correctly characterized), and depth (implicit constraints and structural nuances recognized). This module specifically tests the agent's ability to bridge the formulation gap---extracting mathematical structure from business language.

\subsubsection{Module 2: Formal Modeling}
\label{subsec:mod2}

\textit{Input}: Structured problem specification (from Module 1 or provided directly for partial evaluation).\\
\textit{Output}: Complete mathematical formulation including sets, parameters, decision variables with domains, objective function, and all constraints.\\
\textit{Assessment}: Model correctness (does the formulation capture the intended problem?), model completeness (are all constraints included?), and model parsimony (Occam's razor---are there redundant variables or constraints?). Expert-verified solutions serve as the reference standard.

\subsubsection{Module 3: Algorithm Design \& Implementation}
\label{subsec:mod3}

\textit{Input}: Mathematical formulation (from Module 2 or provided directly).\\
\textit{Output}: Executable code implementing the chosen solution algorithm.\\
\textit{Assessment}: Solution feasibility (verified against reference constraints), solution quality (optimality gap relative to reference solution), code quality (readability, modularity, robustness), and scalability (performance across small/medium/large instances). The multi-scale testing is essential: as MIPLIB-NL demonstrated, methods that succeed on small instances often fail catastrophically at scale.

\subsubsection{Module 4: Report Generation}
\label{subsec:mod4}

\textit{Input}: Solution output from Module 3 along with the original problem description.\\
\textit{Output}: A solution report explaining the modeling approach, algorithm choice, solution results, and business insights.\\
\textit{Assessment}: Completeness (covers all pipeline stages from problem understanding to results), clarity (explanations are accessible to non-expert stakeholders), and reproducibility (reported results can be reproduced by running the submitted code).

\subsection{Cross-Module Consistency Metrics}
\label{subsec:consistency}

The most novel aspect of our evaluation architecture is the systematic assessment of consistency between adjacent modules. We define three consistency metrics:

\paragraph{Understanding--Model Consistency ($\mathcal{C}_{12}$).}
Does the mathematical model (Module 2 output) faithfully formalize the problem understanding (Module 1 output)? This metric detects cases where an agent correctly identifies the problem structure but then formulates a different (typically simpler, template-based) model. Assessment combines automated structural comparison with expert review.

\paragraph{Model--Code Consistency ($\mathcal{C}_{23}$).}
Does the code (Module 3 output) faithfully implement the stated mathematical model (Module 2 output)? This is a critical check that catches implementation errors, constraint omissions, and objective function mismatches. Following the methodology of MIPLIB-NL~\citep{miplbnl2026}, we employ both automated constraint-structure comparison and expert code review.

\paragraph{Code--Report Consistency ($\mathcal{C}_{34}$).}
Does the report (Module 4 output) accurately describe the actual implementation and results (Module 3 output)? This metric detects ``hallucinated reports''---documents that describe an idealized version of the solution rather than what was actually implemented and computed.

\subsection{Partial Module Evaluation}
\label{subsec:partial}

The modular architecture supports flexible evaluation configurations. Researchers may evaluate any contiguous subsequence of modules:

\begin{itemize}[leftmargin=*]
    \item \textbf{Modules 1--2}: Tests understanding and formulation capability (comparable to ORQA + NL4Opt scope, but with business-language inputs).
    \item \textbf{Modules 2--3}: Tests formulation-to-implementation capability (comparable to MIPLIB-NL scope, but with consistency checking).
    \item \textbf{Modules 3--4}: Tests implementation and reporting capability.
    \item \textbf{Modules 1--3}: Tests the core technical pipeline without reporting.
    \item \textbf{Modules 1--4}: Full pipeline evaluation.
\end{itemize}

For partial evaluation, the modules outside the evaluation window are provided with ground-truth inputs (e.g., for evaluating Modules 2--3, Module 2 receives the expert-verified problem specification as input). Cross-module consistency is evaluated for all adjacent pairs within the window.

This flexibility is important because different research questions call for different evaluation scopes. A team developing a better formulation engine needs Modules 1--2 evaluation; a team building a code generation agent needs Modules 2--3; a team building a full R\&D agent needs Modules 1--4.

\section{Benchmark Tasks}
\label{sec:tasks}

\ours{} comprises twelve benchmark tasks drawn from industrial optimization scenarios. Each task is designed to satisfy all four dimensions of Problem Formulation Validity (Section~\ref{subsec:pf_validity}) and includes an expert-verified reference model for automated solution validation.

\subsection{Task Design Principles}

All tasks adhere to the following design principles in Section~\ref{subsec:pf_validity}:

\begin{enumerate}[leftmargin=*]
    \item \textbf{Authentic business framing}: Problem descriptions are written entirely in the language and terminology of the relevant industry domain. In particular, we deliberately avoid using academic optimization terminology such as ``linear programming'' or ``mixed-integer model,'' so that agents must infer the problem structure from the business context rather than from explicit mathematical cues.
    \item \textbf{Anti-template traps}: Each task contains at least one structural feature that is specifically designed to cause standard optimization templates to fail. These traps ensure that agents cannot succeed by simply recognizing a problem category and applying a textbook formulation.
    \item \textbf{Multi-scale instances}: Each task provides instances at three different scales---small, medium, and large---so that we can test both the correctness of an agent's formulation and its scalability and robustness when problem size increases.
    \item \textbf{Reference model validation}: Each task includes an expert-verified reference model whose optimal or near-optimal solutions are known in advance. This reference model enables fully automated checking of both solution feasibility (whether all constraints are satisfied) and solution quality (how close the agent's objective value is to the best known solution).
    \item \textbf{Diverse problem types}: The tasks collectively span a wide range of optimization methodologies---from integer programming to stochastic optimization to robust optimization---in order to prevent any single specialization from dominating benchmark performance.
\end{enumerate}

\subsection{Task Overview}

Table~\ref{tab:tasks} provides an overview of the benchmark by listing one representative example task for each problem type. It is important to note that for each problem type shown in the table, we have prepared 4--5 distinct problem instances, each situated within a different business scenario. This design ensures that strong performance on a single instance cannot be attributed to memorization or overfitting to a specific business context.

\begin{table}[h]
\centering
\caption{Overview of \ours{} benchmark tasks. }
\label{tab:tasks}
\resizebox{\textwidth}{!}{
\begin{tabular}{p{4cm}p{4cm}p{8cm}}
\toprule
\textbf{Problem Type} & \textbf{Task Name} & \textbf{Key Anti-Template Feature} \\
\midrule
\small
 Integer Programming & Custom Roll-Material Fulfillment Scheduling & Online roll selection, limited workshop buffer, and deadline-aware batching make static cutting templates fail. \\
 \midrule
 MDP & Drone Medical Package Delivery & Stochastic order arrivals, battery limits, deadlines, and location-dependent actions make myopic routing infeasible. \\
 \midrule
 Two-Stage Stochastic Optimization & Regional Distribution Network Planning & First-stage facility and inventory decisions must hedge against scenario-dependent demand and second-stage recourse costs. \\
  \midrule
 Combinatorial Bandit / Dynamic Assortment Optimization & Dynamic Fashion Assortment Selection & Bayesian demand learning creates an exploration-exploitation tradeoff under weekly shelf-capacity constraints. \\
  \midrule
 Quantitative Finance Optimization & Robust Investment Strategy & Smooth distributional uncertainty and risk-sensitive loss prevent direct use of standard historical Markowitz optimization. \\
  \midrule
 Robust Optimization & Heterogeneous Computing Resource Scheduling & Resource heterogeneity, uncertain task duration, GPU startup costs, and shared network bottlenecks break deterministic scheduling templates. \\
  \midrule
 Choice-Based Revenue Management / MNL Bandit & Cinema Scheduling with Observable Demand & MNL preference learning is coupled with legal showtime spacing and mandatory golden-time scheduling constraints. \\
  \midrule
 Non-Smooth Optimization & City-Scale Data Center Energy Coordination & Peak-demand max costs, battery charge-discharge logic, and cross-time workload shifting create non-smooth coupled decisions. \\
  \midrule
 Future Information Prediction & Fresh Food Store Replenishment & The task mixes decision-time forecasts with outcomes observed only after ordering, so using realized sales, spoilage, or complaints creates look-ahead leakage. \\
  \midrule
 Semi-definite Programming & Project Team Partitioning & Organizational balance, seniority coverage, and mentorship binding constraints break vanilla Max-Cut SDP relaxation. \\
  \midrule
 Second-Order Cone Programming & Enterprise Annuity Portfolio Optimization & Regulatory allocation rules and portfolio risk constraints prevent naive return-maximizing allocation. \\
  \midrule
 Multi-Stage Stochastic Optimization & E-Commerce Trunk Capacity Planning & Markovian demand, spot-market prices, inventory carryover, and emergency recourse couple decisions across multiple stages. \\
\bottomrule
\end{tabular}
}
\end{table}

The tasks collectively cover: three fundamentally different uncertainty handling paradigms (robust optimization, stochastic programming, and MDP); both convex (SOCP) and non-convex problem structures; both smooth and non-smooth objective functions; single-objective and multi-objective formulations; and both static and dynamic (time-dependent) decision contexts. This diversity ensures that benchmark performance reflects broad optimization R\&D capability rather than specialization in a single problem type.

\subsection{Two Illustrative Tasks}

To illustrate our task design philosophy, we briefly describe the traps embedded in two tasks to demonstrate how our task design satisfies the principles.

\subsubsection{Task T1: Custom Roll-Material Fulfillment Scheduling.} 
A custom roll-material processor receives online orders specifying material types and required lengths, which must be fulfilled by cutting from material rolls under strict per-order deadlines, a workshop capacity limit of 3 rolls, and compatibility rules that forbid substitute materials when exact-type rolls are feasible. The algorithm must make real-time decisions about roll selection, transportation scheduling, and order batching without knowledge of future orders, optimizing a lexicographic objective that prioritizes fulfillment rate over substitution count, transportation count, and material waste.

\begin{tcolorbox}[
    title={\textbf{Task T1: Custom Roll-Material Fulfillment Scheduling}},
    colback=gray!5,
    colframe=black!70,
    fonttitle=\small\bfseries,
    fontupper=\small,
    breakable
]

\paragraph{Background.}
In roll-material processing, customer orders are fulfilled by cutting required lengths from raw material rolls. The current ``random roll selection with per-order round-trips'' approach leads to excessive transportation and material waste. The goal is to design an intelligent scheduling algorithm that optimizes roll selection, transportation, and cutting across all incoming orders.

\paragraph{Order Execution Flow.}
Each order specifies a material type, required length, and creation time. Execution proceeds as: (1)~select a roll that can fully satisfy the order (\textbf{no splitting allowed}---one order must be served by exactly one roll); (2)~if the selected roll is in storage, transport it to the workshop (fixed 30~minutes; multiple rolls may be transported simultaneously); (3)~cut immediately upon arrival (cutting time is negligible); (4)~the roll \textbf{may remain} in the workshop for future orders without being returned immediately.

\paragraph{Optimization Objectives (by priority).}
\begin{enumerate}[leftmargin=*, nosep]
    \item Minimize the number of \textbf{unfilled} orders (highest priority).
    \item Minimize \textbf{substitution count} (prefer exact-type rolls over substitutes).
    \item Minimize total \textbf{transportation count} (both storage$\to$workshop and workshop$\to$storage count as one move each).
    \item Minimize material \textbf{scrap waste}.
\end{enumerate}

\paragraph{Key Constraints.}

\textit{Workshop capacity.} The workshop holds at most 3~rolls normally. A temporary 4th roll may be brought in for cutting, but immediately after cutting one roll must be sent back so that the count returns to~3. If fewer than 3~rolls are present, rolls may stay indefinitely.

\textit{Order deadlines.} Orders created between 00:00--11:59 must be completed (cut finished) within \textbf{70~minutes}; orders created between 12:00--23:59 within \textbf{50~minutes}. Orders exceeding their deadline are marked unfilled. The system operates 24 hours with no downtime.

\textit{Online decision-making.} Orders arrive in real time; the algorithm can only see orders that have already arrived and \textbf{cannot foresee future orders}. The algorithm may batch pending orders and decide when to act, but must respect all deadlines.

\textit{Material compatibility rules.} Compatibility relationships are given (e.g., type~A orders may use type~B rolls). However, substitution is allowed \textbf{only when all exact-type rolls are infeasible}---either insufficient remaining length or transportation would exceed the deadline. Substituting merely to save transportation is \textbf{forbidden}.

\textit{Residual handling.} After cutting: remaining length $\leq 2$\,m is scrap; remaining length $> 2$\,m is retained as tail inventory.

\textit{Zero-length orders.} Automatically fulfilled without consuming any roll.

\paragraph{Input/Output.}
Input consists of \texttt{orders.csv} (type, length, creation time), \texttt{rolls.csv} (type, ID, initial length), and \texttt{compatibility.csv} (compatible material pairs). Output includes: assignment records mapping each order to a roll, a transportation log of all moves with timestamps, workshop state snapshots, final roll inventory, and aggregate metrics (total orders, pulls, substitutions, unfilled count, scrap, and tail meters).

\end{tcolorbox}

\paragraph{Why the task is difficult?}

This task is designed to expose a failure mode we term \emph{problem-category anchoring with algorithmic blind spots}. The business description presents a roll-material processing scenario in which orders arrive online, each requesting a specific material type and length to be cut from a rool. The vocabulary---``cutting,'' ``rools,'' ``lengths,'' ``waste minimization''---strongly invites the agent to anchor to the classical Cutting Stock Problem (CSP) and apply column-generation or bin-packing heuristics. However, the true optimization structure of this task is fundamentally different from a CSP, and an agent that commits to this template will produce a severely suboptimal solution.

The critical insight that separates this task from a standard CSP lies in the \emph{temporal dimension} of the problem. In a classical cutting stock formulation, all orders are known in advance and the objective is purely combinatorial: find cutting patterns that minimize material waste. In this task, however, orders arrive in real time with strict per-order deadlines (70~minutes for morning orders, 50~minutes for afternoon orders), and the dominant cost driver is not material waste but \emph{transportation}---each rool movement between storage and the workshop costs 30~minutes and counts toward the objective. The workshop can hold only 3~rools at a time (with a temporary 4th allowed briefly), meaning that rool residency decisions directly determine which future orders can be served without additional transportation. This transforms the problem from a static combinatorial optimization into a \emph{dynamic scheduling problem with inventory positioning under online uncertainty}.

The anti-template trap operates at two levels. At the \emph{formulation level}, an agent anchoring to the CSP template will focus on minimizing cutting waste through optimal pattern selection---the classical CSP objective. But in this task, waste minimization is the \emph{lowest-priority} objective; the hierarchy is: unfilled orders $\succ$ substitution count $\succ$ transportation count $\succ$ waste. A CSP-based formulation entirely misses the transportation and scheduling dimensions that dominate the objective. Moreover, because the CSP assumes all orders are known upfront, it cannot handle the online arrival constraint: the algorithm is explicitly forbidden from observing future orders, making any static batch-optimization approach infeasible for the full problem.

At the \emph{algorithmic level}, the key mechanism that a correct solution must exploit is \emph{order holding} (also called delayed decision-making). The problem statement explicitly permits the algorithm to ``accumulate arrived but unprocessed orders and decide when to act in batch.'' This is a critical degree of freedom: by strategically delaying decisions on pending orders---waiting to see whether additional orders for the same material type or for rools already in the workshop will arrive---the algorithm can dramatically reduce the total number of rool movements. Consider a simple example: if three orders for type~A material arrive within a 20-minute window, an agent that processes them one-by-one will potentially trigger three separate rool transports, whereas an agent that holds the first two orders and processes all three together using a single rool already positioned in the workshop incurs only one transport (or even zero, if the rool was already present). The optimal holding strategy must balance the benefit of batching (fewer transports, better rool utilization) against the risk of deadline violation (holding too long causes orders to become unfilled).

This ``hold-then-batch'' strategy has no counterpart in the classical CSP literature. It requires reasoning about: (1)~the stochastic value of waiting---how likely is it that a compatible order will arrive soon enough to justify holding? (2)~deadline-aware scheduling---how much slack does each pending order have, and can we afford to wait? (3)~workshop inventory positioning---which 3~r'o'o'ls should we keep in the workshop to maximize the probability of serving future orders without transportation? These considerations place the problem squarely in the domain of online scheduling with limited buffers and lookahead constraints, bearing more resemblance to problems in dynamic vehicle dispatching or semiconductor lot scheduling than to cutting stock.

An agent that recognizes only the ``cutting'' aspect of the problem and applies a CSP solver will generate a solution that processes each order immediately upon arrival, selects rools based purely on waste minimization, and triggers a round-trip transport for nearly every order. Such a solution will have correct cutting patterns but catastrophically high transportation counts---often 2--3$\times$ higher than a solution that exploits order holding. Conversely, an agent that correctly identifies the online scheduling structure and implements a hold-and-batch policy with deadline-aware rool positioning will achieve far fewer transports while still satisfying all feasibility constraints. The anti-template nature of this task thus lies in the gap between what the problem \emph{looks like} (a cutting stock problem where the primary challenge is combinatorial pattern optimization) and what it \emph{actually is} (an online scheduling problem where the primary challenge is temporal decision-making about when to act and which rools to keep nearby).

\subsubsection{Task TT2: Enterprise Annuity Portfolio Optimization.} 

A quantitative analyst must construct a portfolio that is robust to distributional shifts in future asset returns, given $n$ historical return samples across $d$ assets. The uncertainty set is defined by three principles: the worst-case distribution must be continuous and smooth, the worst-case evaluation is "softened" via a risk-sensitive expectation controlled by a smoothing parameter $\epsilon$, and distributional deviations are penalized by a quadratic transport cost with budget radius $\rho$. The objective is to minimize a worst-case loss that combines expected negative return with a penalized CVaR term over this ambiguity set.

\begin{tcolorbox}[
    title={\textbf{Task T2: Robust Investment Portfolio Optimization}},
    colback=gray!5,
    colframe=black!70,
    fonttitle=\small\bfseries,
    fontupper=\small,
    breakable
]

\paragraph{Background.}
As a quantitative analyst, your task is to develop a portfolio optimization strategy that is robust to uncertainty in future asset returns. Standard approaches that rely solely on historical data often perform poorly when market conditions shift. To obtain a portfolio that performs well even under adverse market conditions, we need to consider a broader set of possibilities.

\paragraph{Problem Setting.}
Given $n$ historical asset return samples, each being a vector of returns for $d$ assets, your goal is to find the optimal portfolio weights $\boldsymbol{\theta}$ (a $d$-dimensional vector where $\theta_i \geq 0$ and $\sum_i \theta_i = 1$). The strategy should be resilient against potential shifts of the future return distribution away from the historical empirical distribution.

This uncertainty can be defined through a novel approach based on the following principles, designed to yield more realistic and practical solutions:

\begin{enumerate}[leftmargin=*, nosep]
    \item \textbf{Continuity of uncertainty.} The worst-case future return distribution is assumed to be continuous and smooth, rather than concentrated on a few extreme discrete points. This reflects a belief that the underlying data-generating process is non-discrete.

    \item \textbf{Smooth pessimism.} Instead of a hard minimax approach that seeks a single worst-case outcome around each historical observation, the model employs a ``smoothed'' worst-case evaluation. This is achieved by considering a ``probability cloud'' of potential future returns around each historical data point and computing a \textbf{risk-sensitive expectation} over that cloud. The approach is controlled by a smoothing parameter $\epsilon > 0$: it averages over many adverse scenarios rather than focusing solely on the absolute worst one.

    \item \textbf{Geometry-aware cost.} The model recognizes that some distributional shifts are more likely than others. It introduces a transport cost $c(x, y) = \|x - y\|_2^2$ to penalize scenarios where future returns $y$ deviate significantly from historical observations $x$. The overall budget for distributional uncertainty is defined by a radius $\rho$.
\end{enumerate}

\paragraph{Task Requirements.}

\begin{enumerate}[leftmargin=*]
    \item \textbf{Mathematical formulation.} Formulate the portfolio optimization problem as a minimax optimization problem. The objective should minimize the worst-case loss function $L(\boldsymbol{\theta}, z) = -\boldsymbol{\theta}^\top z + \varrho \cdot \text{P-CVaR}_\alpha(-\boldsymbol{\theta}^\top z)$ over the ambiguity set described by the principles above. Provide the primal form of this problem.

    \item \textbf{Dual reformulation and algorithm design.} Derive the dual form of the optimization problem from Task~1. The resulting dual objective may contain a nested expectation with a log-sum-exp structure. Describe an efficient and scalable algorithm to solve this dual problem, particularly when the number of historical samples $n$ and the number of assets $d$ are large. Explain why standard stochastic gradient methods may encounter difficulties, and describe how your proposed algorithm overcomes these challenges (e.g., by handling biased gradients).

    \item \textbf{Simulated data.} To test your model, you may use the data provided in \texttt{data.csv}.
\end{enumerate}

\end{tcolorbox}

\paragraph{Why the task is difficult?}  
This task is designed to expose a failure mode we term \emph{template anchoring through familiar financial language}. The business description presents what appears to be a standard portfolio allocation problem: a quantitative analyst must find optimal portfolio weights $\boldsymbol{\theta}$ that maximize risk-adjusted returns given $n$ historical return samples across $d$ assets. The description uses familiar financial terminology---``portfolio weights,'' ``asset returns,'' ``worst-case loss,'' ``CVaR''---that strongly invites the agent to anchor to the classical Markowitz mean-variance framework or a standard robust optimization template.

The critical trap lies in the three design principles embedded within the problem description, each of which individually appears innocuous but collectively transforms the problem's mathematical structure into something far removed from any textbook formulation. The first principle---\emph{continuity of uncertainty}---requires that the worst-case adversarial distribution be continuous and smooth rather than discrete. An agent anchoring to standard distributionally robust optimization (DRO) with Wasserstein ambiguity sets will typically formulate the inner maximization as a finite-dimensional problem over discrete support points, which directly violates this requirement. The second principle---\emph{smooth pessimism}---introduces a risk-sensitive expectation controlled by a smoothing parameter $\epsilon > 0$ that replaces the hard minimax over each data point with a soft exponential tilting. This is, in mathematical substance, an \emph{entropic regularization} of the worst-case distribution, but the problem description never uses this term; instead, it describes the mechanism through the intuitive metaphor of a ``probability cloud'' with ``averaged adverse scenarios.'' The third principle---\emph{geometry-aware cost}---introduces a quadratic transport cost $c(x,y) = \|x - y\|_2^2$ with a budget radius $\rho$, which defines a Wasserstein-type ball around the empirical distribution. Again, the term ``optimal transport'' never appears in the problem statement.

The anti-template trap operates at two levels. At the \emph{formulation level}, the agent must recognize that the three stated principles, taken together, define a distributionally robust optimization problem with \emph{entropic optimal transport} (EOT) as the divergence measure---a formulation where the ambiguity set consists of all continuous distributions within an entropically regularized Wasserstein ball of radius $\rho$ around the empirical distribution. An agent that anchors to a standard Markowitz template will produce a convex quadratic program that entirely ignores distributional robustness. An agent that recognizes the robust optimization aspect but misses the entropic regularization will formulate a standard Wasserstein DRO problem whose worst-case distribution is discrete (violating Principle~1) and whose inner problem lacks the smooth structure needed for efficient computation. Only an agent that correctly synthesizes all three principles will arrive at the EOT-regularized minimax formulation.

At the \emph{solution methodology level}, a second trap awaits. The dual reformulation of the EOT-regularized problem yields an objective containing a nested log-sum-exp structure---specifically, an empirical average of $\log \mathbb{E}[\exp(\cdot)]$ terms that must themselves be estimated via integration over the ``probability cloud'' around each data point. An agent that recognizes the dual form may still fail at the algorithmic design stage: standard stochastic gradient descent applied to this objective produces \emph{biased} gradient estimates because the logarithm of an expectation does not commute with sampling. The correct algorithmic approach involves either multi-level Monte Carlo estimators to debias the gradients, or---more elegantly---recognizing that the dual structure admits a \emph{Sinkhorn-type iterative scaling algorithm}. The Sinkhorn algorithm exploits the entropic regularization to decompose the optimization into alternating matrix-scaling operations that converge geometrically, avoiding the bias problem entirely. However, the problem description deliberately avoids any mention of ``Sinkhorn,'' ``matrix scaling,'' or ``iterative proportional fitting,'' requiring the agent to independently identify this algorithmic connection from the mathematical structure alone.

In summary, this task layers two distinct anti-template traps: first, business-language framing that invites standard portfolio theory but actually requires an entropic optimal transport formulation; second, a dual problem structure that appears amenable to stochastic gradient methods but in fact demands a specialized Sinkhorn-type algorithm due to the bias introduced by the log-expectation nesting. An agent that succeeds on this task must demonstrate not only the ability to look past familiar financial terminology to identify the correct mathematical structure, but also the deeper algorithmic insight to connect entropic regularization with iterative scaling methods---a connection that is well-established in the optimal transport literature but is never signaled by the problem's business-oriented language.

\section{Experimental Framework}
\label{sec:experiments}

\subsection{System Architecture}

\begin{figure}[h]
    \centering
    \includegraphics[width=0.4\linewidth]{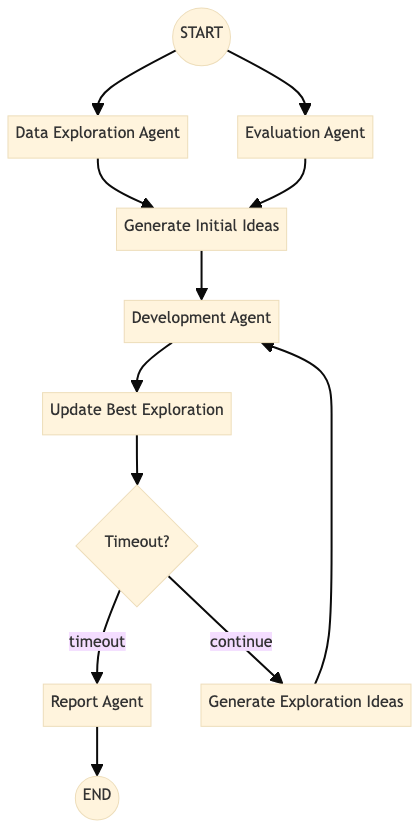}
    \caption{Architecture of the optimization agent pipeline. The system operates in three stages: (1)~parallel initialization, where a Data Exploration Agent retrieves domain-relevant knowledge and an Evaluation Agent constructs assessment criteria; (2)~an iterative development loop that repeatedly generates, implements, and evaluates solution ideas within a configurable time budget, retaining only the best-performing direction at each iteration; and (3)~report generation, where the final solution and its analysis are synthesized into a structured output.}
    \label{fig:rd_agent_pipeline}
\end{figure}

To evaluate agent performance on our benchmark tasks, we develop a dedicated optimization agent built upon the RD-Agent framework~\citep{rdagent2024}. The agent is implemented as an end-to-end system that receives business-language problem descriptions and autonomously produces mathematical formulations, executable solver code, and solution reports. Figure~\ref{fig:rd_agent_pipeline} illustrates the overall pipeline architecture.

The agent operates in three stages. In the first stage, two sub-agents run in parallel: a \emph{Data Exploration Agent} searches for domain knowledge and optimization methods relevant to the given problem, while an \emph{Evaluation Agent} constructs problem-specific evaluation criteria based on its understanding of the task requirements. Users may further guide the evaluation logic through system-prompt instructions. The outputs of both agents are then merged to generate an initial set of solution ideas.

In the second stage, the system enters an \emph{iterative development loop} governed by a configurable time budget. Within each iteration, a \emph{Development Agent} implements the current best idea into executable code using Python-based optimization packages (primarily CVXPY and PuLP for convex formulations). After each implementation attempt, the system evaluates solution quality against the criteria established in Stage~1 and updates the best-performing exploration direction. If time remains, new exploration ideas are generated based on accumulated feedback, and the loop continues. This mechanism ensures that the agent progressively improves its solution by selecting the most promising direction from multiple candidates, rather than committing to a single approach.

In the final stage, a \emph{Report Agent} synthesizes the best solution, its mathematical formulation, implementation details, and performance analysis into a structured report.

Key design decisions include: (1)~domain-specific prompt engineering tailored to operations research problem structures, which guides the agent to reason about objectives, constraints, and solution methodologies in a systematic manner; (2)~solver-agnostic implementation through Python optimization libraries rather than fixed commercial solvers, allowing flexible adaptation to diverse problem types; and (3)~a best-of-$k$ selection mechanism within the time budget that balances exploration breadth against exploitation depth.

\subsection{Evaluation Protocol}

\paragraph{Full-Pipeline Evaluation.}
In the default configuration, agents receive only the business-language task description and data files. They must produce: (1) a structured problem analysis, (2) a mathematical formulation, (3) executable solver code, and (4) a solution report. All four modules and three consistency metrics are evaluated.

\paragraph{Partial-Pipeline Evaluation.}
For targeted capability assessment, researchers can provide ground-truth outputs for upstream modules and evaluate only the downstream modules of interest.

\subsection{Metrics}
\label{subsec:metrics}

Evaluating an LLM's performance on an end-to-end optimization task requires more than a single objective-value comparison. The output spans problem understanding, mathematical modeling, code implementation, and written reporting---each requiring distinct competencies that may succeed or fail independently. To capture this multifaceted nature, \ours{} employs a hierarchical evaluation framework organized into three levels: \emph{module-level metrics} that assess each pipeline stage in isolation, \emph{cross-module consistency metrics} that verify logical coherence between stages, and \emph{aggregate metrics} that enable cross-task and cross-model comparison. This section details how the validity principles established in Section~\ref{sec:solution_validity} are concretely operationalized through rubric-driven evaluation.

\subsubsection{Module-Level Metrics}

Each task is evaluated across four modules, with each module assessed through a structured rubric containing multiple checkpoints organized by severity (Critical / Major / Minor). Table~\ref{tab:metrics} summarizes the metric categories and their constituent assessment dimensions.

\begin{table}[H]
\centering
\small
\caption{Module-level and cross-module metrics in \ours{}.}
\label{tab:metrics}
\begin{tabular}{lll}
\toprule
\textbf{Category} & \textbf{Metric} & \textbf{What It Measures} \\
\midrule
\multirow{3}{*}{\shortstack[l]{Modeling\\(ME)}} & ME.1: Model Correctness & Core mathematical structure and constraint capture \\
& ME.2: Model Parsimony & Whether the formulation avoids unnecessary complexity \\
& ME.3: Model--Code Consistency & Alignment between stated formulation and implementation \\
\midrule
\multirow{3}{*}{\shortstack[l]{Code\\(CE)}} & CE.1: Feasibility \& Correctness & Solution feasibility, constraint satisfaction, optimality \\
& CE.2: Code Quality \& Efficiency & Readability, robustness, and computational scalability \\
& CE.3: Robustness & Performance stability under instance perturbation \\
\midrule
\multirow{4}{*}{\shortstack[l]{Report\\(RE)}} & RE.1: Report Completeness & Coverage of model, method, results, and verification \\
& RE.2: Explanation Clarity & Quality of mathematical and methodological exposition \\
& RE.3: Result Reproducibility & Whether reported numbers match code execution \\
& RE.4: Limitation Acknowledgment & Honest disclosure of method boundaries and assumptions \\
\bottomrule
\end{tabular}
\end{table}

\paragraph{Rubric Design Principles.}
Each metric is operationalized through a set of rubric checkpoints---concrete, binary-verifiable questions about the LLM's output. For example, under ME.1 (Model Correctness), a checkpoint might ask: ``Does the formulation include the SDP lift with $\mathrm{diag}(X)=\mathbf{1}$ and $X \succeq 0$?'' Each checkpoint specifies: (1)~the evidence source (response text, code, or execution result), (2)~the scoring method (automatic via checker, LLM-as-judge, or hybrid), and (3)~the point allocation reflecting its importance within the metric.

Checkpoints are assigned severity levels that determine their weight in the final score:
\begin{itemize}[leftmargin=*]
    \item \textbf{Critical} ($w=3$): Failures indicating fundamental misunderstanding---e.g., misidentifying the problem type, omitting a binding constraint, or using future information unavailable at the decision point. A critical failure caps the maximum achievable score for the entire metric.
    \item \textbf{Major} ($w=2$): Failures that significantly degrade solution quality but do not invalidate the entire approach---e.g., an unnecessarily restrictive relaxation or a solver misconfiguration that prevents convergence at scale.
    \item \textbf{Minor} ($w=1$): Suboptimal but non-fatal choices---e.g., inefficient code structure, incomplete sensitivity analysis, or imprecise variable naming.
\end{itemize}

\paragraph{Anti-Template Alignment.}
A distinctive feature of our rubric design is its explicit coupling with the anti-template traps embedded in each task. For every structural feature designed to cause standard templates to fail (Section~\ref{sec:tasks}), the rubric contains at least one critical checkpoint that directly tests whether the LLM has fallen into the trap. For example, in the Custom Roll-Material Fulfillment Scheduling task (T1), the rubric includes critical checkpoints asking: ``Does the formulation respect online order visibility and deadline-aware roll holding?'' and ``Does the implementation exploit order holding to reduce roll movements?'' An LLM that anchors to a standard CSP template will fail both checkpoints, receiving a low score that accurately reflects its inability to reason beyond familiar problem categories---even if its cutting patterns are individually optimal.

\paragraph{Illustrative Example.}
To make the rubric structure concrete, we briefly illustrate with the Project Team Partitioning task (SDP category). This task requires the LLM to recognize the problem as a constrained Max-Cut, formulate the SDP relaxation ($X = xx^\top$, $\mathrm{diag}(X)=\mathbf{1}$, $X \succeq 0$), and implement constraint-aware rounding. The ME.1 rubric for this task contains seven checkpoints, including: (i)~identifying the problem as Max-Cut rather than generic clustering (0.45 points); (ii)~writing the correct binary variable and cut objective (0.60 points); (iii)~providing the SDP lift with PSD constraint (0.95 points, critical); (iv)~explaining the rank-1 relaxation relationship (0.75 points); (v)~handling all four business constraint types (0.90 points, critical); (vi)~providing constraint-aware rounding and repair (0.85 points); and (vii)~distinguishing SDP relaxation bounds from exact enumeration on small instances (0.50 points). An LLM that solves the 12-person instance via brute-force enumeration can achieve the correct optimal value of 189, but will score at most 2.0/5.0 on ME.1 because it fails the critical SDP-related checkpoints---accurately reflecting that it has not demonstrated the target competency. Complete rubrics for all tasks are provided in Appendix~\ref{app:rubrics}.

\paragraph{Scoring Method Mix.}
Across all tasks, approximately 60\% of checkpoint evaluations are performed automatically (through code execution, constraint checkers, and output parsing), while the remaining 40\% employ an LLM-as-judge protocol with calibrated scoring guidelines. Hybrid checkpoints combine both: automated extraction of evidence followed by semantic judgment on whether the evidence satisfies the criterion.

\subsubsection{Cross-Module Consistency Metrics}

Beyond evaluating each module independently, \ours{} measures whether the LLM's outputs across pipeline stages are logically coherent with one another. These consistency checks directly operationalize the ``Runnable Code $\neq$ Correct Implementation'' and ``Coherent Report $\neq$ Faithful Report'' principles established in Section~\ref{subsec:critical_distinctions}:

\begin{itemize}[leftmargin=*]
    \item $\mathcal{C}_{\text{ME.3}}$\textbf{: Model--Code Consistency.} Verifies that the code implements the same model described in the formulation. The Evaluator Agent performs a ``semantic diff,'' comparing the mathematical structure embedded in the code (e.g., \texttt{Gurobi.addConstr} calls) against the agent's stated formulation. Detects discrepancies such as constraints present in the mathematical model but simplified or omitted in code, or claims of SDP implementation when the code performs enumeration.
    \item $\mathcal{C}_{\text{RE.3}}$\textbf{: Code--Report Consistency.} Verifies that reported results (objective values, constraint satisfaction claims) are reproducible by executing the submitted code. The Evaluator Agent cross-references claims made in the report against actual execution logs and code structure. Detects hallucinated results or stale numbers from prior iterations.
\end{itemize}

These consistency checks are embedded directly within the ME.3 and RE.3 metrics respectively, rather than being computed as separate scores. This design ensures that inconsistencies are penalized in their natural context: a model--code divergence is a modeling failure, while a code--report divergence is a reporting failure.

\subsubsection{Aggregate Metrics}

For cross-task and cross-model comparison, we define the following aggregate metrics:

\begin{itemize}[leftmargin=*]
    \item \textbf{Module Score} $S_m$: The severity-weighted average across all checkpoints within module $m$, normalized to $[0,5]$. For the three-category scheme (ME, CE, RE), the overall module scores are:
    \begin{equation}
    \label{eq:total_score}
        S_{\text{total}} = 0.35 \cdot S_{\text{ME}} + 0.40 \cdot S_{\text{CE}} + 0.25 \cdot S_{\text{RE}}
    \end{equation}
    where $S_{\text{ME}} = \frac{1}{3}(\text{ME.1} + \text{ME.2} + \text{ME.3})$, $S_{\text{CE}} = \frac{1}{2}(\text{CE.1} + \text{CE.2})$ (CE.3 included when available), and $S_{\text{RE}} = \frac{1}{4}(\text{RE.1} + \text{RE.2} + \text{RE.3} + \text{RE.4})$. The weights reflect that implementation correctness is the most consequential capability for practical deployment, followed by modeling quality, and then reporting.

    \item \textbf{Success Rate}: The fraction of tasks where the total score exceeds a threshold $\tau$ (default: $\tau = 3.0$ on the 5-point scale), indicating acceptable overall competence.

    \item \textbf{Anti-Template Score}: The average score on critical checkpoints specifically aligned with anti-template traps, measuring the LLM's resistance to pattern-matching failures.

    \item \textbf{Robustness Score}: For tasks with CE.3 implemented, the ratio of performance on perturbed/scaled instances to performance on base instances, measuring generalization beyond the specific test case. Taking task T3 as an example, Table~\ref{tab:model-methods} presents the solution approaches obtained by each model after iterative refinement, while Table~\ref{tab:combined-robust-summary-all} reports the execution results of each model's generated code across different test instances.

\begin{tcolorbox}[
    title={\textbf{Task T3: Cross-Line Collaboration Optimization}},
    colback=gray!5,
    colframe=black!70,
    fonttitle=\small\bfseries,
    fontupper=\small,
    breakable
]

\paragraph{Background.}
A software company employs $n$ engineers, each belonging to one of three technical disciplines---\emph{frontend}, \emph{backend}, or \emph{data}---and holding one of three seniority levels: \emph{senior}, \emph{mid-level}, or \emph{junior}. The company has decided to split its engineering organization into two parallel product lines (Line~A and Line~B) to accelerate product iteration. Management must simultaneously satisfy several requirements when designing the partition.

The corporate HR system has recorded the pairwise collaboration frequency among all engineers over the past six months (including code reviews, joint debugging sessions, and document co-authoring), yielding an $n \times n$ symmetric weight matrix~$W$. Employees are indexed from~1 to~$n$; for employees~$a$ and~$b$, the entry $W[a{-}1][b{-}1]$ denotes the number of collaborative interactions between them. Management believes that when two frequently collaborating engineers are assigned to \emph{different} product lines, their relationship becomes a ``cross-line bridge,'' providing sustained knowledge flow and technical alignment between the two lines. Accordingly, management seeks to \emph{maximize} the total capacity of such cross-line technical communication channels.

The partition must, however, satisfy the following organizational constraints:

\begin{enumerate}
    \item \textbf{Headcount balance.} Both product lines require sufficient staffing to sustain their respective iteration cadences. Each line must contain at least~3 engineers and at most $\lfloor 0.7n \rfloor$ engineers.

    \item \textbf{Frontend coverage.} Frontend engineers are responsible for user interfaces and interaction design---a critical component of product delivery. Each product line must include at least one frontend engineer to ensure independent frontend development capability.

    \item \textbf{Senior staffing balance.} Senior engineers serve as the core decision-makers on technical matters. To guarantee the quality of architectural design and technical judgment on both lines, each product line must contain at least two senior engineers.

    \item \textbf{Mentor--mentee co-location.} The company operates a mentorship program in which each mentor conducts daily code reviews and provides technical guidance to their mentee. There are $m$ mentor--mentee pairs in total; each pair must be assigned to the \emph{same} product line to preserve continuity of mentorship.
\end{enumerate}

The file \texttt{data1.json} contains the company's employee information. This file serves only as a public example; the final evaluation will use multiple hidden instances conforming to the same schema.

\paragraph{Field specification.}
\begin{verbatim}
employees    : List of employee display names (List[str]);
               the (k-1)-th entry corresponds to employee k.
departments  : List of departments (List[str]);
               the (k-1)-th entry corresponds to employee k.
seniority    : List of seniority levels (List[str]);
               the (k-1)-th entry corresponds to employee k.
mentor_pairs : List of mentor-mentee pairs (List[Tuple[int,int]]);
               uses 1-based employee indices.
W            : n x n symmetric collaboration weight matrix;
               W[a-1][b-1] denotes the collaboration frequency
               between employees a and b.
\end{verbatim}

\paragraph{Output requirements.}
\begin{enumerate}
    \item The submission must be a complete, executable \texttt{.py} file.
    \item The solver must be \emph{general-purpose}. It must not hard-code the number of employees, employee names, mentor--mentee pairs, the dimensions of the weight matrix, or the optimal partition for the provided example.
    \item The program must read the input JSON file path and the output JSON file path from command-line arguments.
    \item The program shall read instance data from the input JSON file and produce an output JSON file with the following structure:
\begin{verbatim}
{
  "line_a": [...],
  "line_b": [...],
  "objective": <number>,
  "method": "...",
  "feasible": true | false
}
\end{verbatim}
    where \texttt{line\_a} and \texttt{line\_b} are lists of 1-based employee indices. The two product lines must jointly cover all employees without overlap and satisfy all hard constraints. The field \texttt{objective} reports the total cross-line collaboration frequency.
    \item The code must generalize to instances of varying scale under the same schema, including changes in~$n$, $m$, department distribution, seniority distribution, mentor--mentee relationships, and the weight matrix.
\end{enumerate}

\end{tcolorbox}

\begin{table}[h]
\centering
\caption{Summary of solution approaches adopted by each model.}
\label{tab:model-methods}
\small
\begin{tabular}{l l }
\toprule
\textbf{Model} & \textbf{Method Type} \\
\midrule
Human
  & SDP Relaxation + GW Rounding + CP-SAT\\
\addlinespace
qwen3.5-plus
  & Multi-start Greedy + Simulated-Annealing-style Local Search\\
\addlinespace
deepseek-v3.2
  & Genetic Algorithm + Local Search \\
\addlinespace
kimi-k2.5
  & Random Restart + Component-level Local Search\\
\addlinespace
qwen3-max
  & OR-Tools CP-SAT Integer Programming \\
\addlinespace
claude-sonnet-4-6\_cloudsway
  & Group-level SA + Individual-level SA \\
\bottomrule
\end{tabular}
\end{table}

\begin{table}[h]
\centering
\caption{Summary of results on four public benchmark instances.}
\label{tab:combined-robust-summary-all}
\small
\begin{tabular}{lrrlrrr}
\toprule
Instance & Employees\# & Mentor Pairs\# & Model & Time(s) & Obj. & Gap \\
\midrule
50 & 50 & 15 & oracle\_sdp\_gw\_cpsat & 151.7 & 3426 & 0 \\
public\_50 & 50 & 15 & qwen3.5-plus & 195.036 & 3426 & 0  \\
public\_50 & 50 & 15 & deepseek-v3.2 & 0.243 & 3426 & 0 \\
public\_50 & 50 & 15 & kimi-k2.5 & 3.38 & 3426 & 0 \\
public\_50 & 50 & 15 & qwen3-max & 30.898 & 3426 & 0 \\
public\_50 & 50 & 15 & claude-sonnet-4-6\_cloudsway & 0.264 & 3341 & 85 \\
public\_80 & 80 & 31 & oracle\_sdp\_gw\_cpsat & 327.9 & 8738 & 0 \\
public\_80 & 80 & 31 & qwen3.5-plus & 60.046 & 8738 & 0 \\
public\_80 & 80 & 31 & deepseek-v3.2 & 0.334 & 8572 & 166 \\
public\_80 & 80 & 31 & kimi-k2.5 & 9.816 & 8738 & 0 \\
public\_80 & 80 & 31 & qwen3-max & 30.607 & 8632 & 106\\
public\_80 & 80 & 31 & claude-sonnet-4-6\_cloudsway & 0.318 & 8069 & 669\\
public\_100 & 100 & 36 & oracle\_sdp\_gw\_cpsat &  & 13161 & 0 \\
public\_100 & 100 & 36 & qwen3.5-plus & 813.885 & 13161 & 0 \\
public\_100 & 100 & 36 & deepseek-v3.2 & 0.702 & 12906 & 255 \\
public\_100 & 100 & 36 & kimi-k2.5 & 23.373 & 13133 & 28  \\
public\_100 & 100 & 36 & qwen3-max & 30.944 & 13058 & 103 \\
public\_100 & 100 & 36 & claude-sonnet-4-6\_cloudsway & 1.743 & 12257 & 904 \\
public\_200 & 200 & 74 & oracle\_sdp\_gw\_cpsat & 324.0 & 52100 & 0 \\
public\_200 & 200 & 74 & qwen3.5-plus & 3065.338 & 52100 & 0 \\
public\_200 & 200 & 74 & deepseek-v3.2 & 2.333 & 51734 & 366 \\
public\_200 & 200 & 74 & kimi-k2.5 & 190.98 & 52052 & 48 \\
public\_200 & 200 & 74 & qwen3-max & 31.016 & 51666 & 434 \\
public\_200 & 200 & 74 & claude-sonnet-4-6\_cloudsway & 3.795 & 51266 & 834 \\
\bottomrule
\end{tabular}
\par\vspace{0.5em}
\footnotesize Note: For the 100-employee instance, multiple solutions attaining the same best-known objective value were found.
\end{table}

\end{itemize}

\subsection{Failure Mode Taxonomy}
\label{sec:failures}

Based on the failure patterns documented in related work and the structural analysis of our benchmark tasks, we identify a taxonomy of failure modes that \ours{} is designed to expose. These failure modes are organized by the pipeline stage where they originate, though their effects often propagate downstream.

\subsubsection{Problem Understanding Failures}
\begin{itemize}
    \item \textbf{Template Anchoring.}
    The agent recognizes superficial keywords (``delivery,'' ``routing,'' ``warehouse'') and anchors to a standard problem template (VRP, TSP, facility location) without verifying that the template matches the actual problem structure. This is the most common failure mode reported in MIPLIB-NL~\citep{miplbnl2026}, where 48.3\% of errors are modeling-level.
    \item \textbf{Constraint Blindness.}
    The agent fails to identify implicit or domain-specific constraints embedded in business language. For example, ``the quality inspection team can only process 200 meters per shift'' defines a capacity constraint that is not signaled by optimization-typical keywords.
    \item \textbf{Distractor Susceptibility.}
    The agent is misled by irrelevant information included in the problem description. Task T7 specifically tests this: the description includes data and context that are irrelevant to the optimization problem, requiring the agent to identify and filter distractors.
\end{itemize}

\subsubsection{Formal Modeling Failures}
\begin{itemize}
    \item \textbf{Constraint Omission.}
    The agent produces a model that captures the objective and some constraints but omits one or more critical constraints, particularly those arising from implicit business requirements or coupling between subsystems.
    \item \textbf{Structural Mismatch.}
    The agent produces a model whose mathematical structure (e.g., linear vs.\ non-convex, deterministic vs.\ stochastic) does not match the problem's true structure. For example, treating an inherently stochastic problem as deterministic, or linearizing a non-convex constraint without justification.
\end{itemize}

\subsubsection{Implementation Failures}
\begin{itemize}
    \item \textbf{Model--Code Divergence.}
    The code implements a different model than the one stated in the formulation. This often manifests as simplified or omitted constraints in the code that are present in the mathematical model, reflecting the agent's difficulty in translating mathematical notation into code.
    \item \textbf{Scalability Collapse.}
    The implementation works on small instances but fails (through timeout, memory exhaustion, or incorrect results) on large instances. This reflects poor algorithm choice or implementation inefficiencies.
\end{itemize}

\subsubsection{Report Failures}

\begin{itemize}
    \item \textbf{Hallucinated Explanations.}
    The report describes modeling decisions, constraint handling, or solution properties that do not match the actual implementation. This is analogous to the hallucination problem in general LLM outputs but applied specifically to technical documentation.
    \item \textbf{Inconsistent Results.}
    The report presents results (objective values, constraint satisfaction claims, performance metrics) that cannot be reproduced by running the submitted code.
\end{itemize}

\subsection{Results}
\label{subsec:results}

\begin{figure}[t]
    \centering
    \includegraphics[width=\linewidth]{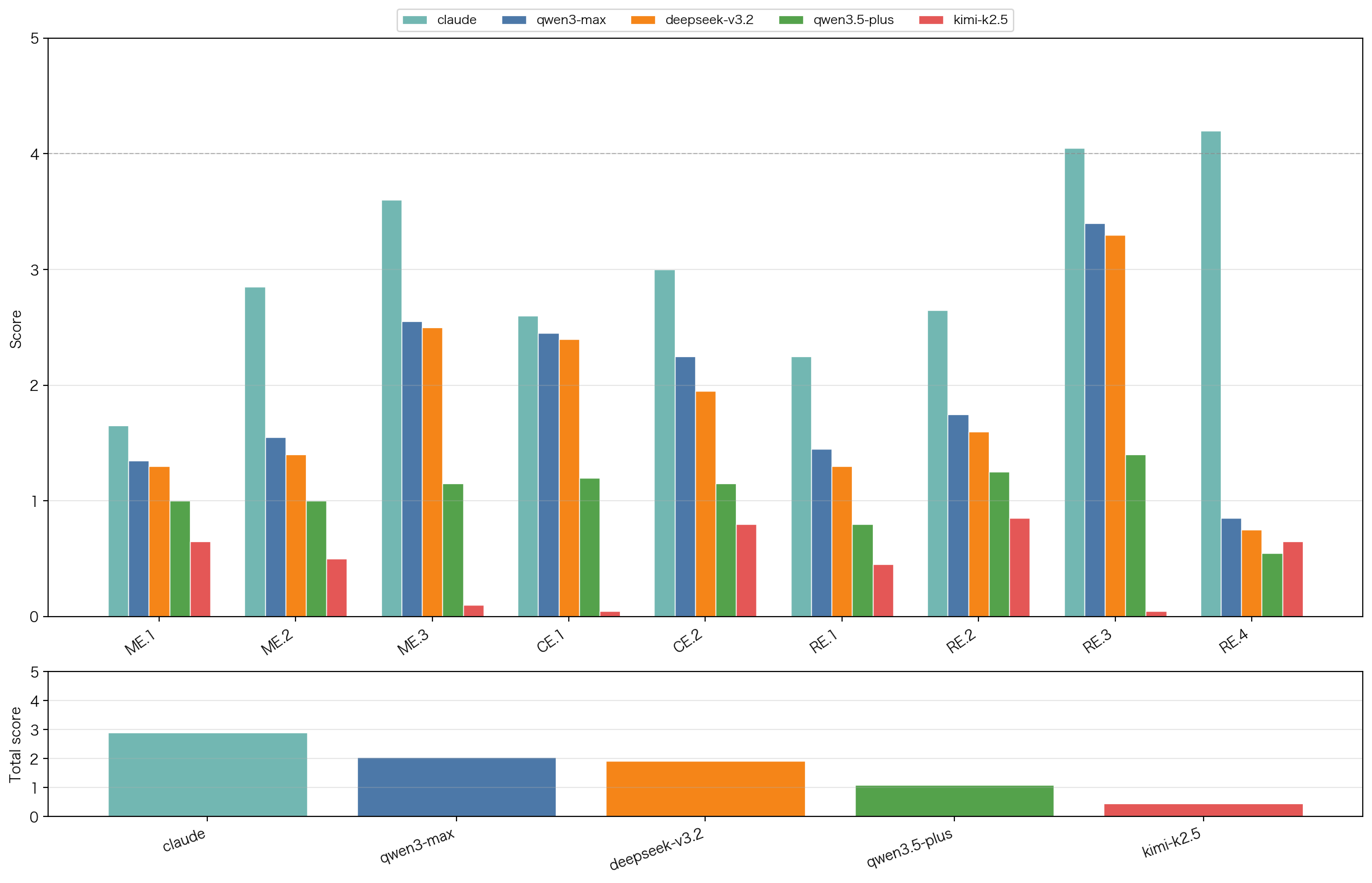}
    \caption{Evaluation results across five LLMs on \ours{}. \textbf{Top}: Average scores on individual metrics (ME.1--ME.3, CE.1--CE.2, RE.1--RE.4) across all benchmark tasks. \textbf{Bottom}: Aggregate total scores computed via Eq.~\eqref{eq:total_score}. All scores are on a 0--5 scale.}
    \label{fig:results}
\end{figure}

We evaluate five frontier LLMs on the full \ours{} benchmark using the pipeline described above: Claude (Sonnet 4), Qwen3-Max, DeepSeek-V3.2, Qwen3.5-Plus, and Kimi-K2.5. Each model is given the same time budget and evaluated under identical conditions using the automated Evaluator Agent. Figure~\ref{fig:results} presents the per-metric scores averaged across all benchmark tasks, as well as the aggregate total scores.

\paragraph{Overall Performance Ranking.}
Claude achieves the highest aggregate total score ($\approx 2.9$), followed by Qwen3-Max ($\approx 2.0$) and DeepSeek-V3.2 ($\approx 1.9$). Qwen3.5-Plus and Kimi-K2.5 trail significantly with aggregate scores near or below 1.0. Notably, even the best-performing model falls well short of the 5.0 ceiling, confirming that \ours{} tasks present genuine difficulty for current frontier models---none have achieved reliable mastery of the full optimization R\&D pipeline.

\paragraph{Metric-Level Analysis.}
Examining the per-metric breakdown reveals several important patterns. First, all models perform relatively better on ME.3 (Model--Code Consistency) and RE.3 (Result Reproducibility) than on ME.1 (Model Correctness), suggesting that current LLMs are more proficient at faithfully translating their own formulations into code than at producing correct formulations in the first place. This aligns with the ``Template Anchoring'' failure mode: models often implement their (incorrect) template faithfully, yielding high consistency scores but low correctness scores.

Second, the performance gap between Claude and other models is most pronounced on ME.2 (Model Parsimony) and RE.4 (Limitation Acknowledgment). Claude demonstrates a stronger ability to avoid unnecessary model complexity and to honestly disclose the boundaries of its approach---capabilities that require deeper problem understanding rather than mere fluency in optimization syntax.

Third, CE.1 (Feasibility \& Correctness) and CE.2 (Code Quality \& Efficiency) show a relatively tight clustering among the top three models but a sharp drop-off for the bottom two, indicating that code-level competence may represent a threshold capability: models either possess sufficient coding fluency for optimization tasks or they do not.

\paragraph{Anti-Template Trap Sensitivity.}
The low absolute scores across all models on ME.1 (all below 3.0, with the weakest models below 1.0) reflect the effectiveness of our anti-template design. Tasks that superficially resemble standard textbook problems but contain structural traps consistently expose template-anchoring behavior. Even Claude, the strongest model, averages only $\approx 1.65$ on ME.1, indicating that it falls into anti-template traps on a substantial fraction of tasks. This validates the Problem Formulation Validity framework (Section~\ref{subsec:pf_validity}): our tasks successfully discriminate between genuine optimization reasoning and pattern-matching competence.

\paragraph{Report Quality vs.\ Technical Quality.}
An interesting observation is that RE.3 (Result Reproducibility) scores are notably high for Claude and Qwen3-Max relative to their other metrics. This suggests that these models are good at producing reports that faithfully describe their actual outputs---even when those outputs are suboptimal. In contrast, Kimi-K2.5 shows near-zero scores on RE.3, indicating a tendency toward hallucinated or fabricated results in its reports, a severe failure mode that undermines any practical utility of the agent's output.

\paragraph{Implications.}
These results demonstrate that \ours{} provides meaningful differentiation across models and reveals capability profiles that would be invisible to simpler evaluation methods. The multi-dimensional scoring exposes not just ``which model is best'' but \textit{where} each model's capabilities break down---information that is directly actionable for targeted model improvement. The consistently low absolute scores further confirm that optimization R\&D remains a challenging frontier for LLM agents, with substantial room for improvement across all evaluated dimensions.

\section{Discussion and Conclusion}
\label{sec:discussion}

The shift from pre-structured to business-language problem descriptions is not merely an increase in difficulty---it fundamentally changes what the benchmark measures. Pre-structured benchmarks primarily test an agent's ability to translate a known mathematical structure into code, which is essentially a \emph{code generation} task. Business-language benchmarks test the agent's ability to perform \emph{optimization research}: identifying the right mathematical framework for an ambiguous real-world situation, a task that requires domain knowledge, mathematical maturity, and creative modeling insight. Organizations seeking to deploy LLM agents for optimization R\&D need to know whether these agents can handle the formulation gap, not just the implementation gap. Our benchmark directly measures this capability.

Cross-module consistency metrics serve a diagnostic function that no single-point metric can provide. An agent might achieve a reasonable objective value (suggesting competent implementation) while its mathematical model omits a critical constraint that happens not to be binding in the test instance. Without model--code consistency checking, this error is invisible. Similarly, an agent might produce an excellent report that describes a sophisticated modeling approach, while its code implements a much simpler model. Without code--report consistency checking, the evaluation would credit the agent for sophistication it did not actually achieve. These consistency checks also provide actionable feedback for agent developers: a high Module~1 score with low $\mathcal{C}_{12}$ consistency indicates that the agent understands the problem but fails in the transition to formal modeling---a specific capability gap that can guide targeted improvement.

Our modular evaluation approach draws inspiration from both SWE-Bench's execution-based evaluation~\citep{swebench2024} and ALE-Bench's score-based iterative assessment~\citep{alebench2025}, but adapts these paradigms for the unique challenges of optimization R\&D. Like SWE-Bench, we use ground-truth validation (Oracle models) rather than output matching. Like ALE-Bench, we support iterative refinement. Unlike both, we explicitly evaluate intermediate pipeline stages and their consistency, reflecting the multi-step reasoning required for optimization research. The MIPLIB-NL finding that fine-tuned models collapse on industrial-scale problems~\citep{miplbnl2026} underscores the importance of our anti-template design. If benchmark tasks can be solved by template matching, then benchmark performance predicts nothing about real-world capability. Our tasks are designed to ensure that template matching is not merely insufficient but actively counterproductive---an agent that template-matches will produce \emph{worse} results than one that starts from scratch with genuine problem analysis.

We acknowledge several limitations of the current framework. First, the largest instances remain below the $10^6$-variable scale of some MIPLIB instances; incorporating ultra-large-scale stress tests would strengthen scalability evaluation. Second, real-world optimization R\&D often involves iterative dialogue with stakeholders and changing requirements; our current framework supports iterative solution refinement but not dynamic problem evolution. 

In summary, we have presented \ours{}, a benchmark that shifts the evaluation of optimization agents from pre-structured mathematical problems to authentic business-language scenarios requiring end-to-end R\&D capability. Our framework makes three key contributions: business-grounded task design with anti-template traps that prevent pattern matching; a modular evaluation architecture that independently assesses problem understanding, formal modeling, implementation, and reporting while enforcing cross-module consistency; and the \orac{} bi-level validity framework that ensures both task quality and scoring integrity. The fundamental insight driving \ours{} is that the hardest part of optimization R\&D is not solving a well-posed mathematical program---it is arriving at the correct formulation from ambiguous, domain-specific business requirements. By directly measuring this capability, along with the downstream skills of implementation and communication, \ours{} provides a more faithful assessment of whether LLM agents can serve as genuine optimization R\&D partners. We release the benchmark tasks, Oracle models, evaluation tools, and \orac{} checklist to the community, and invite contributions of new tasks that satisfy the Problem Formulation Validity criteria.

\bibliographystyle{plainnat}
\bibliography{references}

@inproceedings{swebench2024,
  author    = {Carlos E. Jimenez and John Yang and Alexander Wettig and Shunyu Yao and Kexin Pei and Ofir Press and Karthik Narasimhan},
  title     = {{SWE-bench: Can Language Models Resolve Real-World GitHub Issues?}},
  booktitle = {Proceedings of ICLR},
  year      = {2024},
}

@inproceedings{alebench2025,
  author    = {Yuki Imajuku and Kohki Horie and Yoichi Iwata and Kensho Aoki and Naohiro Takahashi and Takuya Akiba},
  title     = {{ALE-Bench: A Benchmark for Long-Horizon Objective-Driven Algorithm Engineering}},
  booktitle = {Proceedings of NeurIPS},
  year      = {2025},
}

@unpublished{clbench2026,
  author    = {Shihan Dou and others},
  title     = {{CL-Bench: A Benchmark for Context Learning}},
  note      = {Preprint},
  year      = {2026},
}

@InProceedings{rebench2024,
  title = 	 {{RE}-Bench: Evaluating Frontier {AI} R\&D Capabilities of Language Model Agents against Human Experts},
  author =       {Wijk, Hjalmar and Lin, Tao Roa and Becker, Joel and Jawhar, Sami and Parikh, Neev and Broadley, Thomas and Chan, Lawrence and Chen, Michael and Clymer, Joshua M and Dhyani, Jai and Ericheva, Elena and Garcia, Katharyn and Goodrich, Brian and Jurkovic, Nikola and Kinniment, Megan and Lajko, Aron and Nix, Seraphina and Koba Sato, Lucas Jun and Saunders, William and Taran, Maksym and West, Ben and Barnes, Elizabeth},
  booktitle = 	 {Proceedings of the 42nd International Conference on Machine Learning},
  pages = 	 {66772--66832},
  year = 	 {2025},
  volume = 	 {267},
  series = 	 {Proceedings of Machine Learning Research},
  month = 	 {13--19 Jul},
  publisher =    {PMLR},
}

@unpublished{mlebench2024,
  author    = {Jun Shern Chan and others},
  title     = {{MLE-Bench: Evaluating Machine Learning Agents on Machine Learning Engineering}},
  note      = {Preprint},
  year      = {2024},
}

@inproceedings{orqa2025,
  author    = {Mahdi Mostajabdaveh and others},
  title     = {{Evaluating LLM Reasoning in the Operations Research Domain with ORQA}},
  booktitle = {Proceedings of AAAI},
  year      = {2025},
}

@unpublished{miplbnl2026,
  author    = {Zhong Li and Hongliang Lu and Tao Wei and others},
  title     = {{Constructing Industrial-Scale Optimization Modeling Benchmark}},
  note      = {Preprint},
  year      = {2026},
}

@misc{optmath2025,
      title={OptMATH: A Scalable Bidirectional Data Synthesis Framework for Optimization Modeling}, 
      author={Hongliang Lu and Zhonglin Xie and Yaoyu Wu and Can Ren and Yuxuan Chen and Zaiwen Wen},
      year={2025},
      eprint={2502.11102},
      archivePrefix={arXiv},
      primaryClass={cs.AI},
}

@unpublished{optimus2023,
  author    = {Ali AhmadiTeshnizi and Wenzhi Gao and Madeleine Udell},
  title     = {{OptiMUS: Optimization Modeling Using MIP Solvers and Large Language Models}},
  note      = {Preprint},
  year      = {2023},
}

@inproceedings{optibench2025,
  author    = {Zhicheng Yang and Yiwei Wang and Yinya Huang and others},
  title     = {{OptiBench Meets ReSocratic: Measure and Improve LLMs for Optimization Modeling}},
  booktitle = {Proceedings of ICLR},
  year      = {2025},
}

@unpublished{talebi2025,
  author    = {Amirreza Talebi},
  title     = {{Large Language Model-Based Automatic Formulation for Stochastic Optimization Models}},
  note      = {Preprint},
  year      = {2025},
}

@unpublished{pengautomatic2025,
  author    = {Mingming Peng and Zhendong Chen and Jie Yang and others},
  title     = {{Automatic MILP Model Construction for Multi-Robot Task Allocation and Scheduling Based on Large Language Models}},
  note      = {Preprint},
  year      = {2025},
}

@unpublished{cetinkaya2025,
  author    = {Ibrahim Oguz Cetinkaya and I. Esra Buyuktahtakin and Parshin Shojaee and Chandan K. Reddy},
  title     = {{Discovering Heuristics with Large Language Models for Mixed-Integer Programs: Single-Machine Scheduling}},
  note      = {Preprint},
  year      = {2025},
}

@inproceedings{milpevolve2025,
  author    = {Sirui Li and Janardhan Kulkarni and Ishai Menache and Cathy Wu and Beibin Li},
  title     = {{Towards Foundation Models for Mixed Integer Linear Programming}},
  booktitle = {Proceedings of ICLR},
  year      = {2025},
}

@misc{loop2026,
      title={LOOP: A Plug-and-Play Neuro-Symbolic Framework for Enhancing Planning in Autonomous Systems}, 
      author={Ronit Virwani and Ruchika Suryawanshi},
      year={2025},
      eprint={2508.13371},
      archivePrefix={arXiv},
      primaryClass={cs.AI},
}

@unpublished{mathv2024,
  author    = {Ke Wang and Junting Pan and Weikang Shi and others},
  title     = {{Measuring Multimodal Mathematical Reasoning with MATH-Vision Dataset}},
  note      = {Preprint},
  year      = {2024},
}

@inproceedings{milpreduction2025,
author = {Li, Yixuan and Chen, Can and Li, Jiajun and Duan, Jiahui and Han, Xiongwei and Zhong, Tao and Chau, Vincent and Wu, Weiwei and Wang, Wanyuan},
title = {Fast and interpretable mixed-integer linear program solving by learning model reduction},
year = {2025},
isbn = {978-1-57735-897-8},
publisher = {AAAI Press},
booktitle = {Proceedings of the Thirty-Ninth AAAI Conference on Artificial Intelligence and Thirty-Seventh Conference on Innovative Applications of Artificial Intelligence and Fifteenth Symposium on Educational Advances in Artificial Intelligence},
articleno = {3018},
numpages = {9},
series = {AAAI'25/IAAI'25/EAAI'25}
}

@unpublished{knuth2026,
  author    = {Donald E. Knuth},
  title     = {{Claude's Cycles}},
  note      = {Preprint},
  year      = {2026},
}

@article{miplib2017,
  author    = {Ambros Gleixner and Gregor Hendel and Gerald Gamrath and Tobias Achterberg and Michael Bastubbe and Timo Berthold and others},
  title     = {{MIPLIB 2017: Data-Driven Compilation of the 6th Mixed-Integer Programming Library}},
  journal   = {Mathematical Programming Computation},
  volume    = {13},
  number    = {3},
  pages     = {443--490},
  year      = {2021},
}

@inproceedings{nl4opt2023,
  author    = {Rindranirina Ramamonjison and others},
  title     = {{NL4Opt Competition: Formulating Optimization Problems Based on Their Natural Language Descriptions}},
  booktitle = {Proceedings of NeurIPS Competition Track},
  year      = {2023},
}

@misc{rdagent2024,
  author    = {{Microsoft Research}},
  title     = {{RD-Agent: A Research and Development Agent Framework}},
  howpublished = {\url{https://github.com/microsoft/RD-Agent}},
  year      = {2024},
}

@inproceedings{bigcodebench2025,
  author    = {Terry Yue Zhuo and Minh Chien Vu and Jenny Chim and Han Hu and others},
  title     = {{BigCodeBench: Benchmarking Code Generation with Diverse Function Calls and Complex Instructions}},
  booktitle = {Proceedings of ICLR},
  year      = {2025},
}

@inproceedings{livecodebench2025,
  author    = {Naman Jain and King Han and others},
  title     = {{LiveCodeBench: Holistic and Contamination-Free Evaluation of Large Language Models for Code}},
  booktitle = {Proceedings of ICLR},
  year      = {2025},
}

@inproceedings{dsbench2025,
  author    = {Liqiang Jing and Zhehui Huang and Xiaoyang Wang and Wenlin Yao and Wenhao Yu and Kaixin Ma and Hongming Zhang and Xiaodan Du and Dong Yu},
  title     = {{DSBench: How Far Are Data Science Agents from Becoming Data Science Experts?}},
  booktitle = {Proceedings of ICLR},
  year      = {2025},
}

@unpublished{frontiermath2024,
  author    = {Elliot Glazer and Ege Erdil and others},
  title     = {{FrontierMath: A Benchmark for Evaluating Advanced Mathematical Reasoning in AI}},
  institution = {Epoch AI},
  note      = {Preprint},
  year      = {2024},
}

@inproceedings{omnimath2025,
  author    = {Bofei Gao and Feifan Song and Zhe Yang and others},
  title     = {{Omni-MATH: A Universal Olympiad Level Mathematic Benchmark for Large Language Models}},
  booktitle = {Proceedings of ICLR},
  year      = {2025},
}

@inproceedings{snell2025,
  author    = {Charlie Snell and Jaehoon Lee and Kelvin Xu and Aviral Kumar},
  title     = {{Scaling LLM Test-Time Compute Optimally Can Be More Effective than Scaling Model Parameters}},
  booktitle = {Proceedings of ICLR (Oral)},
  year      = {2025},
}

@unpublished{limipsynth2023,
  author    = {Qingyang Li and Lele Zhang and Vicky Mak-Hau},
  title     = {{Synthesizing Mixed-Integer Linear Programming Models from Natural Language Descriptions}},
  note      = {Preprint (arXiv:2311.15271)},
  year      = {2023},
}

@unpublished{optengine2026,
  author    = {Yitian Chen and Cheng Cheng and Yinan Sun and Zi Ling and Dongdong Ge},
  title     = {{OPT-Engine: Benchmarking the Limits of LLMs in Optimization Modeling via Complexity Scaling}},
  note      = {Preprint (arXiv:2601.19924)},
  year      = {2026},
}

@inproceedings{cobench2025,
  author    = {Wei Sun and others},
  title     = {{CO-Bench: Benchmarking Language Model Agents in Algorithm Search for Combinatorial Optimization}},
  booktitle = {Proceedings of AAAI},
  year      = {2026},
}

@inproceedings{astorga2025,
  title={Autoformulation of Mathematical Optimization Models Using {LLM}s},
  author={Nicol{\'a}s Astorga and Tennison Liu and Yuanzhang Xiao and Mihaela van der Schaar},
  booktitle={Forty-second International Conference on Machine Learning},
  year={2025},
}

@article{llmopt2025,
  publtype={informal},
  author={Caigao Jiang and Xiang Shu and Hong Qian and Xingyu Lu and Jun Zhou and Aimin Zhou and Yang Yu},
  title={LLMOPT: Learning to Define and Solve General Optimization Problems from Scratch},
  year={2024},
  cdate={1704067200000},
  journal={CoRR},
  volume={abs/2410.13213},
}

@inproceedings{optitree2025,
  title={OptiTree: Hierarchical Thoughts Generation with Tree Search for {LLM} Optimization Modeling},
  author={Haoyang Liu and Jie Wang and Yuyang Cai and Xiongwei Han and Yufei Kuang and Jianye HAO},
  booktitle={The Thirty-ninth Annual Conference on Neural Information Processing Systems},
  year={2026},
}

@unpublished{orllmagent2025,
  author    = {Bowen Zhang and Pei Luo and Guang Yang and Boon-Hee Soong and Chau Yuen},
  title     = {{OR-LLM-Agent: Automating Modeling and Solving of Operations Research Optimization Problems with Reasoning LLM}},
  note      = {Preprint (arXiv:2503.10009)},
  year      = {2025},
}

@article{orlm2025,
  author    = {Chengyue Huang and Zhixing Tang and Siyuan Hu and Ruoqing Zhang and Yimeng Mao and Yong Chen},
  title     = {{ORLM: A Customizable Framework in Training Large Models for Automated Optimization Modeling}},
  journal   = {Operations Research (INFORMS)},
  year      = {2025},
}

@article{funsearch2024,
  author    = {Bernardino Romera-Paredes and Mohammadamin Barekatain and Alexander Novikov and Matej Balog and others},
  title     = {{Mathematical Discoveries from Program Search with Large Language Models}},
  journal   = {Nature},
  volume    = {625},
  pages     = {468--475},
  year      = {2024},
}

@inproceedings{reevo2024,
  author    = {Haoran Ye and Jiarui Wang and Zhiguang Cao and others},
  title     = {{ReEvo: Large Language Models as Hyper-Heuristics with Reflective Evolution}},
  booktitle = {Proceedings of NeurIPS},
  year      = {2024},
}

@inproceedings{xiaosurvey2025,
  author    = {Zhuohan Xiao and Liunian Xu and Xuan Han and Xianfeng Fu and Jiaxuan Zhu and Tao Zhong and Yushi Wang and Dongbo Zhang and others},
  title     = {{A Survey of Optimization Modeling Meets LLMs: Progress and Future Directions}},
  booktitle = {Proceedings of IJCAI (Survey Track)},
  year      = {2025},
}

@article{liusurveyalgo2025,
  publtype={informal},
  author={Fei Liu and Yiming Yao and Ping Guo and Zhiyuan Yang and Zhe Zhao and Xi Lin and Xialiang Tong and Mingxuan Yuan and Zhichao Lu and Zhenkun Wang and Qingfu Zhang},
  title={A Systematic Survey on Large Language Models for Algorithm Design},
  year={2024},
  cdate={1704067200000},
  journal={CoRR},
  volume={abs/2410.14716},
}

\appendix
\newpage

\section{Illustrative Benchmark Task I: Dynamic Assortment Planning}
\label{app:example1}

\subsection{Problem Overview}

This project aims to design a 20-week new-product launch strategy for a fast-fashion brand, with the goal of identifying high-potential items using minimal pilot inventory and maximizing overall profit. The core challenge lies in balancing the exploration-exploitation trade-off (demand-unknown items vs. high-potential items), while accounting for substitution effects when 30 out of 100 SKUs are launched each week: when a particular item is out of stock, consumers switch to substitute items, creating demand spillover. The operational framework is based on Bayesian dynamic updating, using a Poisson demand model ($\lambda \sim \text{Gamma}$) and a substitution matrix $\mathbf{A}$ (describing inter-SKU substitution probabilities) to compute ``effective demand'' in real time, with weekly closed-loop iteration for assortment decisions. An algorithmic prototype is required to process \texttt{products.csv} (containing price/cost/prior demand) and \texttt{substitution.csv} data, output cumulative profit and Regret metrics, and validate the effectiveness of substitution-effect modeling through comparative experiments. Implementation must balance demand forecasting, substitution relationship capture, and inventory cost control.

\subsection{System Prompt}

As the Chief Digital Operations Officer of a fast-fashion brand, you are responsible for the season-long new-product launch strategy, with the goal of \textbf{``using as little pilot inventory as possible to quickly identify high-potential items and maximize overall profit.''}

Building on the existing ``small-batch testing, rapid iteration'' framework, we further consider \textbf{inter-product substitution relationships}, which adds new complexity to assortment decisions.

\paragraph{\textbf{Operational Rhythm and Constraints}}

\begin{itemize}
    \item Season length: 20 weeks ($T = 20$)
    \item Weekly shelf capacity: 30 SKUs; candidate pool fixed at 100 SKUs
    \item Weekly closed loop: pull actual sales at weekend $\rightarrow$ immediately update model $\rightarrow$ decide next week's keep/replace SKUs
\end{itemize}

\paragraph{\textbf{Base Demand --- Invisible Weekly Sales Potential}}

\begin{itemize}
    \item Each SKU has an invisible ``weekly latent demand'' $\lambda_s$ (units/week).
    \item Actual sales $n_s$ are influenced by noise such as foot traffic, discounts, and weather, modeled by a Poisson distribution:
    \begin{equation}
        n_s \sim \text{Poisson}(\lambda_s)
    \end{equation}
\end{itemize}

\paragraph{\textbf{Prior Belief}}

\begin{equation}
    \lambda_s \sim \text{Gamma}(m_s, \alpha_s)
\end{equation}

\begin{itemize}
    \item $m_s$ approximates historical cumulative sales; $\alpha_s$ approximates cumulative observed weeks
\end{itemize}

\paragraph{\textbf{Substitution Effect Matrix $\mathbf{A}$}}

\begin{itemize}
    \item A new $S \times S$ matrix $\mathbf{A}$ is introduced, where element $A_{s_1,s_2}$ denotes ``the probability that a customer switches to SKU $s_2$ when SKU $s_1$ is not on the shelf'' ($\sum_{s_2 \neq s_1} A_{s_1,s_2} \leq 1$).
    \item Only one substitution occurs: if $s_1$ is not on the shelf and a customer switches to $s_2$, but $s_2$ is also not on the shelf, this portion of demand is lost directly.
    \item \textbf{Effective Demand}
    
    For the actually launched assortment $\mathcal{A}_t$ this week, the effective weekly demand for SKU $s \in \mathcal{A}_t$ is
    \begin{equation}
        \lambda_s^{\text{eff}} = \lambda_s + \sum_{k \notin \mathcal{A}_t} A_{k,s} \cdot \lambda_k
    \end{equation}
    
    Then sales are still generated via a Poisson process:
    \begin{equation}
        n_s \sim \text{Poisson}(\lambda_s^{\text{eff}})
    \end{equation}
\end{itemize}

\paragraph{\textbf{Rolling Learning}}

\begin{itemize}
    \item Only launched SKUs are observed for sales and updated via Bayes:
    \begin{equation}
        m_s \leftarrow m_s + n_s, \quad \alpha_s \leftarrow \alpha_s + 1
    \end{equation}
    \item The substitution inflow is simultaneously treated as true sales of $s$ (because the customer did indeed purchase it).
\end{itemize}

\paragraph{\textbf{Data}}

\begin{itemize}
    \item \texttt{products.csv} provides basic business attributes of each product:
    \begin{itemize}
        \item \texttt{SKU\_No}: product number, a unique ID for each product.
        \item \texttt{price}: product selling price.
        \item \texttt{cost}: product cost, used to calculate profit.
        \item \texttt{m}: prior demand intensity, indicating how popular this product is likely to be.
        \item \texttt{alpha}: prior observation intensity, indicating how reliable this demand information is.
    \end{itemize}
    \item \texttt{substitution.csv}: represents the substitution matrix among products (row = out-of-stock item, column = substitute item; diagonal elements are 0)
\end{itemize}

\paragraph{\textbf{Core KPI}}

Maximize season-end cumulative profit.

\paragraph{\textbf{Task Requirements}}

\begin{enumerate}
    \item \textbf{Assortment Optimization Approach}
    \begin{itemize}
        \item How to simultaneously handle, within the weekly ``100 choose 30'' cycle:
        \begin{itemize}
            \item[a)] exploration-exploitation trade-off
            \item[b)] demand spillover and intra-category competition caused by inter-SKU substitution
        \end{itemize}
    \end{itemize}
    
    \item \textbf{Runnable Code Prototype}
    \begin{itemize}
        \item Read \texttt{products.csv} and $\mathbf{A}$, simulate 20 weeks of operations, output cumulative profit and Regret
    \end{itemize}
    
    \item \textbf{Report}
    \begin{itemize}
        \item Solution rationale, algorithm details, experimental controls (including a no-substitution baseline), and business implementability analysis
    \end{itemize}
\end{enumerate}

\subsection{Rubrics}

\begin{table}[H]
\centering
\caption{Rubrics for evaluating solution quality.}
\small
\begin{tabular}{|c|p{2.0cm}|p{3.6cm}|p{3.6cm}|p{2.8cm}|c|}
\hline
\textbf{No.} & \textbf{Dimension} & \textbf{Check Point} & \textbf{Pass Criteria} & \textbf{Fail Criteria} & \textbf{Severity} \\
\hline
1 & Problem Type & Recognize combinatorial nature of substitution effects & Explicitly state that effective demand depends on the launched assortment & Still treat SKUs independently & Critical \\
\hline
2 & Problem Type & Recognize Bayesian learning as the main thread & Still maintain Gamma-Poisson / TS as the main thread & Abandon dynamic learning & Major \\
\hline
3 & Objective & Quarterly cumulative profit objective is clear & Profit objective is consistent with substitution effects & Objective function is confused & Major \\
\hline
4 & Constraints & ``100 choose 30'' constraint is complete & Operational constraints are correct & Basic constraints are missing & Critical \\
\hline
5 & Modeling & Effective demand formula is handled reasonably & Able to implement or approximate $\lambda_{\text{eff}}$ logic & Substitution matrix logic is severely wrong & Critical \\
\hline
6 & Modeling & Substitution matrix dimensions and coverage are reasonable & At least correctly handle matrix scale and indexing & Dimension mismatch causes large-scale SKU distortion & Critical \\
\hline
7 & Learning & TS / dynamic decision-making truly exists & Not just static sorting & Pure greedy or fake TS & Critical \\
\hline
8 & Code Quality & Code is runnable & Has end-to-end executable implementation & No code or Hello World & Critical \\
\hline
9 & Result Quality & Profit results and solution quality are credible & Output is self-consistent with problem scale & Numbers are groundless or severely distorted & Major \\
\hline
10 & Evaluation Quality & Able to recognize evaluation gaming risks & Has anti-gaming awareness & Completely no security thinking & Minor \\
\hline
\end{tabular}
\label{tab:rubrics}
\end{table}

\subsection{Reference Answer}

\begin{algorithm}[H]
\caption{Reference Algorithm for Dynamic Assortment Planning}
\label{alg:oracle}
\begin{algorithmic}[1]
\Require Candidate set size $S$, shelf capacity $N$, horizon $T$
\Require Unit profit $r[s]$, substitution matrix $Q[i][s]$
\Require Gamma prior parameters $m[s], c[s]$
\State \textbf{Precompute:} offline coefficient table $z_t$ (e.g., from Table~2)
\Function{Index}{$s, t, m, c$}
    \State $\mu \gets m[s]/c[s]$ \Comment{Expected demand rate}
    \State $\sigma^2 \gets m[s]/c[s]^2$ \Comment{Demand rate variance}
    \State $\text{nb\_var} \gets \mu \cdot (1 + 1/c[s])$ \Comment{Neg.-binomial variance}
    \State $\text{idx} \gets r[s] \cdot \bigl( \mu + z_t[t] \cdot \sqrt{\sigma^2 / \text{nb\_var}} \bigr)$
    \State \Return idx \Comment{Desirability index}
\EndFunction
\For{$t = T, T-1, \dots, 1$}
    \State Compute base index: $\text{base\_idx}[s] \gets \textsc{Index}(s,t,m,c), \; \forall s$
    \State \textbf{Assortment optimization:} choose $U \subseteq \{1..S\}, |U|=N$ to maximize
    \begin{equation*}
        \text{OBJ}(U) = \sum_{s \in U} \text{base\_idx}[s] + \sum_{i \notin U} \sum_{s \in U} r[s] \cdot Q[i,s] \cdot \mathbb{E}[\lambda_i]
    \end{equation*}
    \Statex \hspace{\algorithmicindent}\textit{Method:} enumerate if $S$ small; otherwise greedy / local search / MILP
    \State Generate sales $n[s]$:
    \Statex \hspace{\algorithmicindent} $d_i \sim \text{Poisson}(\lambda_i)$; if $i \in U$: $n[i] \gets d_i$ else substitute to $s \in U$ w.p.~$Q[i,s]$
    \For{each $s \in U$} \Comment{Posterior update (estimate original sales)}
        \State $\tilde{n} \gets n[s] \cdot \mu_s / \bigl( \mu_s + \sum_{i \neq s} (1-\mathbf{1}_{i \in U}) \cdot Q[i,s] \cdot \mu_i \bigr)$
        \State $m[s] \gets m[s] + \tilde{n}$; \; $c[s] \gets c[s] + 1$
    \EndFor
\EndFor
\Ensure Optimal (or approximate) assortment $U_t$ for each period and cumulative revenue
\end{algorithmic}
\end{algorithm}

\newpage
\section{Detailed Evaluation Rubric: Project Team Partitioning (SDP)}
\label{app:rubrics}

This appendix provides the complete evaluation rubric for one representative task---Project Team Partitioning (Semi-Definite Programming category)---to illustrate the rubric design methodology described in Section~\ref{subsec:metrics}. Rubrics for all other tasks follow the same structural template and are available in the benchmark repository.

\subsection*{Problem Synopsis}

The task requires partitioning 12 team members into two product lines (A and B) to maximize cross-line collaboration, subject to organizational constraints (team size balance, frontend coverage, senior engineer coverage, and mentor-mentee binding). The underlying mathematical structure is a constrained Max-Cut problem amenable to SDP relaxation.

\subsection*{Scoring Rules}

\begin{itemize}[leftmargin=*]
    \item Each metric is scored on a 5-point scale via checkpoint summation.
    \item The total score is computed as: $S_{\text{total}} = 0.35 \cdot S_{\text{ME}} + 0.40 \cdot S_{\text{CE}} + 0.25 \cdot S_{\text{RE}}$.
    \item If an LLM's response lacks SDP lift / PSD constraint / rank relaxation / rounding, ME.1 is capped at 2.0/5 and RE.2 is capped at 2.6/5.
    \item If the code is purely enumeration-based, CE.1 is capped at 2.6/5 even if it reproduces the optimal value of 189.
    \item CE.3 (Robustness) is marked \texttt{NOT\_IMPLEMENTED} for this task and excluded from the total.
\end{itemize}

\subsection*{ME.1: Model Correctness (5 points)}

Assesses whether the LLM formulates the problem as a constrained Max-Cut with proper SDP relaxation, rather than treating it as a generic clustering or enumeration problem.

\begin{table}[H]
\centering
\small
\begin{tabular}{clllr}
\toprule
\textbf{ID} & \textbf{Checkpoint} & \textbf{Evidence} & \textbf{Method} & \textbf{Pts} \\
\midrule
ME1-R1 & Identifies as constrained Max-Cut & text & LLM Judge & 0.45 \\
ME1-R2 & Correct binary variables and cut objective & text/code & Hybrid & 0.60 \\
ME1-R3 & SDP lift: $X=xx^\top$, $\mathrm{diag}(X)=\mathbf{1}$, $X \succeq 0$ & text/code & Hybrid & 0.95 \\
ME1-R4 & Explains rank-1 vs.\ SDP relaxation relationship & text & LLM Judge & 0.75 \\
ME1-R5 & Handles all four business constraint types & text/code & Hybrid & 0.90 \\
ME1-R6 & Constraint-aware rounding/repair/checker flow & text/code & Hybrid & 0.85 \\
ME1-R7 & Distinguishes SDP bound from exact enumeration & text & LLM Judge & 0.50 \\
\bottomrule
\end{tabular}
\end{table}

\subsection*{ME.2: Model Parsimony (5 points)}

Assesses whether the formulation uses an appropriate minimal structure for the SDP benchmark, without conflating evaluation logic or organizational consulting advice with the core model.

\begin{table}[H]
\centering
\small
\begin{tabular}{clllr}
\toprule
\textbf{ID} & \textbf{Checkpoint} & \textbf{Evidence} & \textbf{Method} & \textbf{Pts} \\
\midrule
ME2-R1 & Clear modeling pipeline: Max-Cut $\to$ SDP $\to$ rounding $\to$ check & text & LLM Judge & 1.15 \\
ME2-R2 & Enumeration positioned as auxiliary oracle, not the SDP method & text/code & Hybrid & 1.10 \\
ME2-R3 & Not over-reliant on brittle bit-ops or irrelevant evaluation scripts & text/code & Hybrid & 0.85 \\
ME2-R4 & Report focuses on math model and algorithm, not generic advice & text & LLM Judge & 0.90 \\
ME2-R5 & Acknowledges small-scale enumerability vs.\ large-scale SDP need & text & LLM Judge & 1.00 \\
\bottomrule
\end{tabular}
\end{table}

\subsection*{ME.3: Model--Code Consistency (5 points)}

Assesses whether the textual formulation and the submitted code implement the same mathematical model.

\begin{table}[H]
\centering
\small
\begin{tabular}{clllr}
\toprule
\textbf{ID} & \textbf{Checkpoint} & \textbf{Evidence} & \textbf{Method} & \textbf{Pts} \\
\midrule
ME3-R1 & Text objective matches code optimization target & text+code & Hybrid & 0.85 \\
ME3-R2 & Text constraints match code constraint checks & text+code & Hybrid & 1.00 \\
ME3-R3 & If text claims SDP, code implements SDP (not enumeration) & text+code & Hybrid & 1.15 \\
ME3-R4 & Reported solution and objective match code output & text+exec & Hybrid & 1.00 \\
ME3-R5 & No ``evaluator code + SDP report'' contradiction & text+code & Hybrid & 1.00 \\
\bottomrule
\end{tabular}
\end{table}

\subsection*{CE.1: Code Feasibility \& Correctness (5 points)}

Assesses whether the code functions as a solver (not merely an evaluator) and produces a feasible, high-quality solution.

\begin{table}[H]
\centering
\small
\begin{tabular}{clllr}
\toprule
\textbf{ID} & \textbf{Checkpoint} & \textbf{Evidence} & \textbf{Method} & \textbf{Pts} \\
\midrule
CE1-R1 & Extractable, executable solver code provided & code/exec & Automatic & 0.50 \\
CE1-R2 & Implements SDP/PSD/rounding (at least two core steps) & code & Hybrid & 2.00 \\
CE1-R3 & Output covers all 12 members with no duplicates/omissions & exec/checker & Automatic & 0.45 \\
CE1-R4 & Output satisfies all four hard constraints & checker & Automatic & 0.60 \\
CE1-R5 & Objective value verifiable; 189 = full marks on this sub-item & checker/oracle & Automatic & 0.50 \\
CE1-R6 & Self-checks, exception handling, and clear entry point & code/exec & Hybrid & 0.95 \\
\bottomrule
\end{tabular}
\end{table}

\subsection*{CE.2: Code Quality \& Efficiency (5 points)}

Assesses code robustness, readability, and reusability beyond the single test instance.

\begin{table}[H]
\centering
\small
\begin{tabular}{clllr}
\toprule
\textbf{ID} & \textbf{Checkpoint} & \textbf{Evidence} & \textbf{Method} & \textbf{Pts} \\
\midrule
CE2-R1 & Clear separation of data, objective, constraints, solve, output & code & LLM Judge & 0.90 \\
CE2-R2 & Correct indexing, names, mentor pairs, and attribute stats & code/checker & Hybrid & 1.00 \\
CE2-R3 & Pipeline-compatible output interface & code/output & Hybrid & 0.75 \\
CE2-R4 & Complexity discussion: enumeration vs.\ SDP/ILP at scale & text/code & LLM Judge & 1.00 \\
CE2-R5 & Reproducibility: dependencies, seeds, failure handling & code/text & Hybrid & 0.85 \\
CE2-R6 & Does not sacrifice problem semantics for evaluation-script compliance & text/code & Hybrid & 0.50 \\
\bottomrule
\end{tabular}
\end{table}

\subsection*{RE.1: Report Completeness (5 points)}

\begin{table}[H]
\centering
\small
\begin{tabular}{clllr}
\toprule
\textbf{ID} & \textbf{Checkpoint} & \textbf{Evidence} & \textbf{Method} & \textbf{Pts} \\
\midrule
RE1-R1 & Restates objective, data, and four hard constraints & text & LLM Judge & 0.60 \\
RE1-R2 & Provides SDP lift, PSD constraint, rank relaxation math & text & LLM Judge & 1.30 \\
RE1-R3 & Describes rounding/repair/checker flow (not just enumeration result) & text & LLM Judge & 1.00 \\
RE1-R4 & Reports final partition, objective, and constraint verification & text/checker & Hybrid & 0.90 \\
RE1-R5 & Clear reproduction instructions: entry point, dependencies, files & text/code & Hybrid & 0.70 \\
RE1-R6 & Distinguishes ``instance optimal 189'' from ``SDP relaxation quality'' & text & LLM Judge & 0.50 \\
\bottomrule
\end{tabular}
\end{table}

\subsection*{RE.2: Explanation Clarity (5 points)}

\begin{table}[H]
\centering
\small
\begin{tabular}{clllr}
\toprule
\textbf{ID} & \textbf{Checkpoint} & \textbf{Evidence} & \textbf{Method} & \textbf{Pts} \\
\midrule
RE2-R1 & Explains cross-line collaboration as graph cut & text & LLM Judge & 0.85 \\
RE2-R2 & Explains PSD matrix, rank-1, SDP relaxation, and rounding & text & LLM Judge & 1.45 \\
RE2-R3 & Explains mathematical/business meaning of four constraint types & text & LLM Judge & 0.90 \\
RE2-R4 & Clarifies roles of enumeration oracle, SDP approximation, and checker & text & LLM Judge & 0.90 \\
RE2-R5 & Well-organized prose without generic filler & text & LLM Judge & 0.90 \\
\bottomrule
\end{tabular}
\end{table}

\subsection*{RE.3: Result Reproducibility (5 points)}

\begin{table}[H]
\centering
\small
\begin{tabular}{clllr}
\toprule
\textbf{ID} & \textbf{Checkpoint} & \textbf{Evidence} & \textbf{Method} & \textbf{Pts} \\
\midrule
RE3-R1 & Clear entry point, dependencies, and input data specified & text/code & Hybrid & 0.70 \\
RE3-R2 & Final partition reconstructible from code output or report & exec/text & Automatic & 0.90 \\
RE3-R3 & Objective value matches independent checker computation & checker & Automatic & 0.80 \\
RE3-R4 & Constraint satisfaction claims match checker results & checker & Automatic & 0.80 \\
RE3-R5 & No confusion among optimal value, baseline, and evaluator discrepancies & text+checker & Hybrid & 1.80 \\
\bottomrule
\end{tabular}
\end{table}

\subsection*{RE.4: Limitation Acknowledgment (5 points)}

\begin{table}[H]
\centering
\small
\begin{tabular}{clllr}
\toprule
\textbf{ID} & \textbf{Checkpoint} & \textbf{Evidence} & \textbf{Method} & \textbf{Pts} \\
\midrule
RE4-R1 & Acknowledges combinatorial hardness; SDP is relaxation/approximation & text & LLM Judge & 0.90 \\
RE4-R2 & Admits enumeration only feasible for 12-person instance & text & LLM Judge & 1.00 \\
RE4-R3 & Honestly reports unimplemented SDP, runtime failures, or inconsistencies & text+code & Hybrid & 1.20 \\
RE4-R4 & Identifies and explains data/baseline/evaluator discrepancies & text & LLM Judge & 1.00 \\
RE4-R5 & Does not overclaim ``global optimal,'' ``full marks,'' or ``production-ready'' & text & LLM Judge & 0.90 \\
\bottomrule
\end{tabular}
\end{table}

\subsection*{Expected Failure Modes}

The following failure patterns are commonly observed on this task and are specifically targeted by the rubric's critical checkpoints:

\begin{enumerate}[leftmargin=*]
    \item Treating the SDP task as a pure enumeration problem, achieving the correct answer (189) without demonstrating the target competency.
    \item Providing only a partition and objective value without explaining SDP lift, PSD constraints, rank relaxation, or rounding.
    \item Code reproduces 189 via enumeration while the report claims SDP completion.
    \item Bit-manipulation errors in mentor-pair checking, yielding an objective value lower than reported.
    \item Top-level code is an evaluator (scores a given partition) rather than a solver (finds a partition).
    \item Confusion among the problem statement's mentor pairs, the evaluation script's pairs, and stale baseline values.
\end{enumerate}

\subsection*{Cross-Model Comparison (Illustrative)}

Table~\ref{tab:sdp_comparison} shows scores for five LLMs evaluated with this rubric, demonstrating how the framework distinguishes between models that achieve the same optimal value but differ in their demonstrated methodology.

\begin{table}[H]
\centering
\small
\caption{Rubric scores for the Project Team Partitioning task across five LLMs. CE.3 is excluded (not implemented for this task).}
\label{tab:sdp_comparison}
\begin{tabular}{lcccccccccc|c}
\toprule
\textbf{Model} & \textbf{ME.1} & \textbf{ME.2} & \textbf{ME.3} & \textbf{CE.1} & \textbf{CE.2} & \textbf{RE.1} & \textbf{RE.2} & \textbf{RE.3} & \textbf{RE.4} & \textbf{Total} \\
\midrule
Qwen3-Max & 3.15 & 4.15 & 4.50 & 5.00 & 4.65 & 3.85 & 3.40 & 4.90 & 2.30 & 4.21 \\
DeepSeek-V3.2 & 2.75 & 4.20 & 5.00 & 4.75 & 4.45 & 3.65 & 3.45 & 4.80 & 2.00 & 4.10 \\
Claude & 2.75 & 3.90 & 4.25 & 5.00 & 3.90 & 3.35 & 2.45 & 4.05 & 1.30 & 3.75 \\
Qwen3.5-Plus & 2.65 & 3.00 & 3.00 & 3.80 & 2.60 & 3.15 & 3.25 & 2.85 & 1.80 & 2.98 \\
Kimi-K2.5 & 2.80 & 2.10 & 0.30 & 0.25 & 2.60 & 3.05 & 2.75 & 0.30 & 2.10 & 1.69 \\
\bottomrule
\end{tabular}
\end{table}

Key observations: (1)~Qwen3-Max, DeepSeek-V3.2, and Claude all reproduce the optimal value 189, yet score differently because the rubric distinguishes enumeration-based solving from SDP-based methodology; (2)~Qwen3.5-Plus's bit-manipulation bug causes actual output of 175 (not the claimed 189), detected by CE.1 and RE.3 automated checkers; (3)~Kimi-K2.5's top-level code is an evaluator rather than a solver, causing near-zero scores on implementation and reproducibility metrics despite reasonable textual exposition.

\section{Task Examples}

\subsection{Semidefinite Programming: Portfolio Risk Optimization}

\subsubsection{Problem Statement}

\textbf{Background:}

A wealth management company manages a portfolio consisting of $N$ stocks. The company aims to reduce the overall risk of the portfolio without sacrificing expected returns. Traditional risk measurement methods assume stock returns follow a normal distribution, but historical data shows markets frequently exhibit ``fat tail'' phenomena---extreme events occur far more often than normal distribution predicts.

The company has collected daily return data from the past 5 years and found that certain stocks exhibit abnormal co-movement during market downturns. The CFO noticed that traditional variance-covariance matrices cannot capture this tail correlation, causing the portfolio's actual losses during market turbulence to far exceed expectations.

Additionally, the company faces practical constraints:
\begin{itemize}
    \item Regulatory requirements stipulate that no single stock's weight may exceed 15\% of total portfolio value
    \item To maintain sufficient diversification, at least 20 different stocks must be held
    \item Certain sectors (such as technology and energy) cannot exceed 30\% combined weight to avoid sector concentration risk
    \item The company wants to ensure that in the worst 5\% of market scenarios, daily portfolio losses do not exceed 2\% of total assets
\end{itemize}

\textbf{Problem Description:}

Please help the company design a portfolio allocation scheme. You need to consider the covariance structure of returns, particularly how to accurately model correlations between stocks, especially during periods of market stress. At the same time, you need to ensure all regulatory and internal risk management requirements are satisfied.

\textbf{Known Data:}
\begin{itemize}
    \item The size of $N$ is given in \texttt{numbers.json}
    \item Expected daily return vector for $N$ stocks is given in \texttt{revenue.csv}
    \item Historical daily return data from the past 5 years is given in matrix \texttt{hist\_revenue.csv}
    \item Current market volatility is at historically high levels, with tail risk significantly elevated
\end{itemize}

\textbf{Decision Objective:}

Find a portfolio weight allocation such that under a risk measure that accurately reflects tail correlation, overall risk is minimized while ensuring expected returns are not below a certain target level and all constraints are satisfied.

\subsubsection{Mathematical Modeling Hints}

This problem requires simultaneous handling of the following complexities:
\begin{enumerate}
    \item \textbf{Risk Measure Selection:} Traditional variance may underestimate tail risk. Consider using robust risk measurement methods based on semidefinite matrices.
    \item \textbf{Correlation Structure Modeling:} Correlations between stocks may change with market states. A method capable of capturing conditional correlations is needed.
    \item \textbf{Constraint Coupling:} The diversification constraint (at least 20 stocks) interacts with the single-stock weight upper bound (15\%), creating a complex combinatorial structure in the feasible solution space.
    \item \textbf{Tail Risk Constraint:} The loss limit in the worst 5\% scenario is a conditional constraint involving the distribution tail, requiring special treatment.
    \item \textbf{Market State Impact:} The current high-volatility environment means historical correlations may not accurately predict future risk, necessitating market state modeling.
\end{enumerate}

The core challenge of this problem lies in: How to optimize the allocation through a risk measure that accurately captures tail correlation while maintaining the portfolio's expected return and satisfying all practical constraints. Directly applying the standard mean-variance optimization framework may fail because the standard framework cannot handle tail correlation and conditional risk constraints.

\subsubsection{Problem Formulation Validity Evaluation}

\textbf{Problem Summary:}

A wealth management company must optimize a 50-stock portfolio to minimize tail-aware risk while satisfying regulatory constraints (single stock limit, diversification minimum, sector caps) and a conditional value-at-risk constraint, requiring robust covariance estimation that captures extreme market co-movements.

\textbf{Detailed Scores:}

\begin{table}[h]
\centering
\begin{tabular}{|p{4.5cm}|c|p{6cm}|}
\hline
\textbf{Criterion} & \textbf{Score} & \textbf{Evidence} \\
\hline
TD.1 Business Semantic Authenticity & 1.0 & Uses pure business language: ``wealth management company'', ``portfolio'', ``fat tail phenomena'', ``CFO'', ``regulatory requirements'', ``sector concentration risk''. No optimization jargon appears. \\
\hline
TD.2 Anti-Template Design & 1.0 & Multiple traps: standard mean-variance fails; CVaR constraint requires special handling; diversification constraint creates combinatorial structure; high-volatility market state requires robust estimation. \\
\hline
TD.3 Multi-Constraint Coupling & 1.0 & Strong interactions: 15\% limit + 20-stock minimum creates complex feasible region; sector caps interact with individual limits; CVaR constraint couples all decisions through tail correlation. \\
\hline
TD.4 Anti-Pattern-Matching & 1.0 & Keywords deliberately misleading: ``portfolio optimization'' suggests mean-variance but problem explicitly rejects this; ``correlation'' requires conditional/tail correlation, not Pearson. \\
\hline
\end{tabular}
\end{table}

\textbf{Overall Score: 1.00 / 1.00}

\textbf{Assessment:}

This problem falls into the \textbf{Excellent} category (0.75-1.00), representing a high-quality benchmark task that forces genuine modeling effort. The problem successfully disguises its underlying mathematical structure (a semidefinite program for robust portfolio optimization with CVaR constraints) behind authentic business language and multiple anti-template elements. The combination of tail risk modeling, complex constraint interactions, and the explicit rejection of standard approaches makes template matching impossible. This is an ideal benchmark problem for testing whether agents can derive appropriate mathematical formulations from business requirements rather than relying on pattern recognition.

\textbf{Strengths:}
\begin{enumerate}
    \item \textbf{Authentic Financial Context:} Realistic wealth management scenario with genuine concerns about tail risk, regulatory compliance, and sector concentration. References to ``CFO observations'' and ``regulatory requirements'' ground the problem in real business practice.
    \item \textbf{Sophisticated Anti-Template Design:} Explicitly warns that standard approaches will fail (``traditional variance-covariance matrices cannot capture this tail correlation''), forcing recognition of need for advanced techniques like semidefinite programming.
    \item \textbf{Complex Constraint Interactions:} Interplay between single-stock limits, diversification minimums, sector caps, and tail risk constraints creates rich optimization landscape.
    \item \textbf{State-Dependent Parameters:} High volatility market state introduces time-varying risk requiring adaptive modeling.
\end{enumerate}

\textbf{Areas for Improvement:}
\begin{enumerate}
    \item \textbf{Data Specification:} Clarify whether covariance estimation from $R$ is required or if covariance matrix is provided.
    \item \textbf{Target Return:} Add concrete minimum return target value for self-contained problem statement.
\end{enumerate}

\textbf{Recommendations:}

This problem is already of excellent quality for benchmark purposes. To further enhance, consider adding transaction costs or turnover constraints, multiple market regimes with different correlation structures, or explicit robustness requirements against covariance estimation uncertainty.

\subsubsection{SDP Mathematical Formulation Details}

\textbf{Why This is a Semidefinite Programming (SDP) Problem}

The above portfolio problem requires solving a robust portfolio optimization problem, with core challenges in:

\textbf{1. Covariance Matrix Uncertainty Modeling}

Traditional mean-variance optimization uses a fixed covariance matrix $\Sigma$, but this problem explicitly states that historical covariance cannot capture tail correlations. Therefore, an uncertainty set needs to be introduced:

$$\mathcal{U} = \left\{ \Sigma \succeq 0 : \Sigma_{ij} \in [\underline{\Sigma}_{ij}, \overline{\Sigma}_{ij}], \|\Sigma - \hat{\Sigma}\|_F \leq \rho \right\}$$

where $\hat{\Sigma}$ is the sample covariance matrix and $\rho$ is the uncertainty radius. This set requires $\Sigma$ to be a positive semidefinite matrix.

\textbf{2. Robust Risk Minimization}

The risk under worst-case covariance is:

$$\max_{\Sigma \in \mathcal{U}} w^T \Sigma w$$

This maximization problem can be transformed into SDP constraints. Introducing auxiliary variable $t$:

$$\min_{w, t} \quad t$$

$$\text{s.t.} \quad \begin{bmatrix} t & w^T \\ w & \Sigma^{-1} \end{bmatrix} \succeq 0, \quad \forall \Sigma \in \mathcal{U}$$

\textbf{3. CVaR Constraint SDP Reformulation}

The Conditional Value-at-Risk (CVaR) constraint that losses in the worst 5\% scenario do not exceed 2\% can be expressed as:

$$\text{CVaR}_{0.95}(-w^T r) \leq 0.02$$

Using Rockafellar-Uryasev's representation theorem, this is equivalent to:

$$\min_{\alpha} \left\{ \alpha + \frac{1}{0.05} \mathbb{E}[(-w^T r - \alpha)^+] \right\} \leq 0.02$$

This can be reformulated using linear constraints and auxiliary variables, which when combined with the semidefinite constraint on the covariance matrix, yields a complete SDP formulation.

\subsection{Two-Stage Stochastic Optimization: Regional Distribution Network Planning}

\subsubsection{Problem Statement}

\textbf{Background:}

An e-commerce company is planning to build warehouses in East China. The company needs to decide warehouse locations and capacities in the first stage, before knowing the exact demand. In the second stage, after demand is realized (under one of three possible scenarios), the company can adjust inventory levels, perform transshipments between warehouses, and arrange replenishment and distribution.

The key challenge is balancing the fixed investment cost of building warehouses against the operational flexibility and recourse costs in different demand scenarios.

\textbf{Known Data:}
\begin{itemize}
    \item Candidate warehouse locations and their construction costs
    \item Three demand scenarios with associated probabilities
    \item Transportation costs between warehouses and customer zones
    \item Inventory holding costs and shortage penalties
    \item Warehouse capacity expansion costs
\end{itemize}

\textbf{Decision Objective:}

Minimize the total expected cost, including first-stage fixed investment and second-stage expected operational costs across all scenarios.

\subsubsection{Problem Formulation Validity Evaluation}

\textbf{Problem Summary:}

An e-commerce company must make facility location and capacity decisions under demand uncertainty, with recourse actions including inventory adjustment, transshipment, and replenishment across three demand scenarios.

\textbf{Detailed Scores:}

\begin{table}[h]
\centering
\begin{tabular}{|p{4.5cm}|c|p{6cm}|}
\hline
\textbf{Criterion} & \textbf{Score} & \textbf{Evidence} \\
\hline
TD.1 Business Semantic Authenticity & 0.9 & Uses authentic logistics terminology: ``warehouse location'', ``capacity planning'', ``transshipment'', ``demand scenarios''. Mathematical notation is minimal and contextual. \\
\hline
TD.2 Anti-Template Design & 0.85 & Standard facility location templates fail due to multi-scenario recourse structure; transshipment decisions create network flow dependencies across scenarios. \\
\hline
TD.3 Multi-Constraint Coupling & 0.9 & First-stage capacity decisions constrain second-stage flexibility; scenario probabilities affect optimal recourse strategies; budget constraints couple location and capacity choices. \\
\hline
TD.4 Anti-Pattern-Matching & 0.85 & ``Warehouse planning'' suggests simple location-allocation, but problem requires two-stage stochastic programming with scenario-dependent recourse. \\
\hline
\end{tabular}
\end{table}

\textbf{Overall Score: 0.88 / 1.00}

\textbf{Assessment:}

This problem represents a high-quality benchmark for two-stage stochastic optimization. It requires agents to recognize the need for scenario-based modeling and understand the interaction between first-stage strategic decisions and second-stage operational adjustments.

\subsection{Multi-Stage Stochastic Optimization: Next-Day Delivery E-commerce Trunk Capacity Planning}

\subsubsection{Problem Statement}

\textbf{Background:}

An e-commerce platform promises ``next-day delivery'' to customers. The platform must make capacity decisions at multiple stages:
\begin{itemize}
    \item \textbf{Stage 1 (Dawn):} Book buy-out trucks with tiered discounts before demand is known
    \item \textbf{Stage 2 (Morning):} Purchase spot market capacity at higher prices or use expensive air freight as recourse
\end{itemize}

The challenge is handling the ``double whammy'' of bad luck: if demand exceeds expectations, the platform faces both higher spot prices and potential service failures.

\textbf{Known Data:}
\begin{itemize}
    \item Tiered pricing structure for buy-out trucks
    \item Spot market price distribution
    \item Air freight emergency cost
    \item Demand distribution forecasts
    \item Service level requirements
\end{itemize}

\textbf{Decision Objective:}

Minimize expected total cost while meeting next-day delivery commitments.

\subsubsection{Problem Formulation Validity Evaluation}

\textbf{Problem Summary:}

An e-commerce platform must make multi-stage capacity decisions under demand uncertainty, with tiered pricing, spot market recourse, and emergency air freight options, requiring careful handling of non-anticipativity constraints and scenario-dependent costs.

\textbf{Detailed Scores:}

\begin{table}[h]
\centering
\begin{tabular}{|p{4.5cm}|c|p{6cm}|}
\hline
\textbf{Criterion} & \textbf{Score} & \textbf{Evidence} \\
\hline
TD.1 Business Semantic Authenticity & 0.95 & Pure logistics language: ``next-day delivery'', ``buy-out trucks'', ``spot market'', ``air freight''. No optimization jargon. \\
\hline
TD.2 Anti-Template Design & 0.9 & Tiered pricing creates non-linear cost structure; ``double whammy'' effect prevents simple newsvendor solutions; multi-stage structure requires dynamic programming. \\
\hline
TD.3 Multi-Constraint Coupling & 0.85 & Service level constraints interact with cost minimization; tiered discounts create threshold effects; emergency options provide recourse but at premium cost. \\
\hline
TD.4 Anti-Pattern-Matching & 0.9 & ``Capacity planning'' suggests simple forecasting, but problem requires multi-stage stochastic programming with non-linear costs. \\
\hline
\end{tabular}
\end{table}

\textbf{Overall Score: 0.90 / 1.00}

\textbf{Assessment:}

This problem effectively tests multi-stage stochastic optimization capabilities. The tiered pricing structure and ``double whammy'' effect create a rich decision landscape that prevents simple template matching.

\subsection{Robust Optimization: Heterogeneous Computing Power Scheduling}

\subsubsection{Problem Statement}

\textbf{Background:}

A cloud computing provider operates heterogeneous computing resources with varying performance characteristics. Computing demands fluctuate unpredictably, and hardware performance may degrade under certain conditions. The provider needs to design resource allocation schemes that remain stable even under worst-case scenarios.

\textbf{Known Data:}
\begin{itemize}
    \item Available computing resources with performance ranges
    \item Demand uncertainty sets
    \item Hardware degradation models
    \item Service level agreements (SLAs)
\end{itemize}

\textbf{Decision Objective:}

Ensure service availability in extreme scenarios while optimizing resource utilization under normal conditions.

\subsubsection{Problem Formulation Validity Evaluation}

\textbf{Problem Summary:}

A cloud provider must allocate heterogeneous computing resources under demand and performance uncertainty, requiring robust optimization techniques to guarantee SLA compliance in worst-case scenarios.

\textbf{Detailed Scores:}

\begin{table}[h]
\centering
\begin{tabular}{|p{4.5cm}|c|p{6cm}|}
\hline
\textbf{Criterion} & \textbf{Score} & \textbf{Evidence} \\
\hline
TD.1 Business Semantic Authenticity & 0.9 & Uses cloud computing terminology: ``heterogeneous resources'', ``performance degradation'', ``SLA'', ``demand fluctuation''. \\
\hline
TD.2 Anti-Template Design & 0.85 & Uncertainty sets for both demand and performance create nested robustness requirements; standard deterministic scheduling fails. \\
\hline
TD.3 Multi-Constraint Coupling & 0.8 & SLA constraints must hold for all uncertainty realizations; resource heterogeneity creates complex feasibility regions. \\
\hline
TD.4 Anti-Pattern-Matching & 0.85 & ``Resource allocation'' suggests simple assignment, but problem requires robust counterpart transformation. \\
\hline
\end{tabular}
\end{table}

\textbf{Overall Score: 0.85 / 1.00}

\subsection{Second-Order Cone Programming: Enterprise Annuity Investment Scheme Design}

\subsubsection{Problem Statement}

\textbf{Background:}

An enterprise needs to design an annuity investment scheme for its employees. The scheme must balance expected returns against risk, with complex asset correlation structures requiring strict control over variance and downside risk.

\textbf{Known Data:}
\begin{itemize}
    \item Available investment assets with expected returns
    \item Covariance matrix of asset returns
    \item Regulatory constraints on asset allocation ratios
    \item Risk tolerance levels
\end{itemize}

\textbf{Decision Objective:}

Maximize expected return under given risk limits, or minimize tracking error subject to asset allocation ratio constraints.

\subsubsection{Problem Formulation Validity Evaluation}

\textbf{Problem Summary:}

An enterprise must design an annuity investment portfolio using second-order cone programming to handle quadratic risk constraints and complex asset correlations while satisfying regulatory allocation requirements.

\textbf{Detailed Scores:}

\begin{table}[h]
\centering
\begin{tabular}{|p{4.5cm}|c|p{6cm}|}
\hline
\textbf{Criterion} & \textbf{Score} & \textbf{Evidence} \\
\hline
TD.1 Business Semantic Authenticity & 0.9 & Uses financial planning language: ``annuity investment'', ``asset allocation'', ``risk tolerance'', ``tracking error''. \\
\hline
TD.2 Anti-Template Design & 0.85 & Quadratic risk constraints require SOCP formulation; standard linear programming cannot handle variance constraints efficiently. \\
\hline
TD.3 Multi-Constraint Coupling & 0.8 & Allocation ratio constraints interact with risk limits; covariance structure creates non-trivial feasible regions. \\
\hline
TD.4 Anti-Pattern-Matching & 0.85 & ``Investment design'' suggests simple diversification, but problem requires convex optimization with quadratic constraints. \\
\hline
\end{tabular}
\end{table}

\textbf{Overall Score: 0.85 / 1.00}

\subsection{Markov Decision Process (MDP): UAV Medical Parcel Delivery}

\subsubsection{Problem Statement}

\textbf{Background:}

A healthcare provider uses unmanned aerial vehicles (UAVs) to deliver medical parcels on a 4×4 grid island. The UAV has limited battery capacity, and orders arrive randomly with strict timeout penalties. The system must decide at each time step whether to fly, pick up packages, deliver, or recharge.

\textbf{Known Data:}
\begin{itemize}
    \item Grid layout and distances
    \item UAV battery consumption rates
    \item Order arrival distributions
    \item Timeout penalty structure
    \item Recharging station locations
\end{itemize}

\textbf{Decision Objective:}

Maximize total expected reward over 20 time steps, balancing flight consumption, waiting losses, and successful delivery bonuses.

\subsubsection{Problem Formulation Validity Evaluation}

\textbf{Problem Summary:}

A UAV delivery system on a grid must make sequential decisions under uncertainty, balancing battery constraints, delivery deadlines, and recharging needs, requiring MDP or reinforcement learning approaches.

\textbf{Detailed Scores:}

\begin{table}[h]
\centering
\begin{tabular}{|p{4.5cm}|c|p{6cm}|}
\hline
\textbf{Criterion} & \textbf{Score} & \textbf{Evidence} \\
\hline
TD.1 Business Semantic Authenticity & 0.9 & Uses logistics terminology: ``UAV delivery'', ``medical parcels'', ``timeout penalties'', ``battery capacity''. \\
\hline
TD.2 Anti-Template Design & 0.9 & State-space explosion prevents brute-force solutions; exploration-exploitation tradeoff requires RL/DP; terminal conditions for battery depletion add complexity. \\
\hline
TD.3 Multi-Constraint Coupling & 0.85 & Battery constraints couple spatial and temporal decisions; timeout penalties create urgency-pressure interactions. \\
\hline
TD.4 Anti-Pattern-Matching & 0.9 & ``Delivery routing'' suggests static optimization, but problem requires sequential decision-making under uncertainty. \\
\hline
\end{tabular}
\end{table}

\textbf{Overall Score: 0.89 / 1.00}

\subsection{Gittins Index / Bandit Algorithms: Multi-Arm Bandit for Product Selection}

\subsubsection{Problem Statement}

\textbf{Background:}

A marketing team faces multiple product options with unknown attractiveness (e.g., marketing channels, suppliers). They must balance trying new options (exploration) and choosing the current best (exploitation) to maximize cumulative rewards.

\textbf{Known Data:}
\begin{itemize}
    \item Number of arms (options)
    \item Prior distributions (Beta-Binomial or Normal-Normal)
    \item Reward observation mechanism
    \item Time horizon
\end{itemize}

\textbf{Decision Objective:}

Maximize cumulative expected reward while minimizing regret relative to the known optimal strategy.

\subsubsection{Problem Formulation Validity Evaluation}

\textbf{Problem Summary:}

A marketing team must solve a multi-arm bandit problem with conjugate priors, requiring Gittins index calculation or Thompson sampling for online learning and regret minimization.

\textbf{Detailed Scores:}

\begin{table}[h]
\centering
\begin{tabular}{|p{4.5cm}|c|p{6cm}|}
\hline
\textbf{Criterion} & \textbf{Score} & \textbf{Evidence} \\
\hline
TD.1 Business Semantic Authenticity & 0.85 & Uses marketing language: ``product selection'', ``marketing channels'', ``supplier evaluation''. \\
\hline
TD.2 Anti-Template Design & 0.9 & Conjugate prior structure requires specific index calculations; standard greedy or $\epsilon$-greedy strategies are suboptimal. \\
\hline
TD.3 Multi-Constraint Coupling & 0.7 & Single-dimensional decision per period, but long-term regret depends on entire history. \\
\hline
TD.4 Anti-Pattern-Matching & 0.85 & ``Product selection'' suggests A/B testing, but problem requires sophisticated bandit algorithms. \\
\hline
\end{tabular}
\end{table}

\textbf{Overall Score: 0.83 / 1.00}

\subsection{Non-Smooth Optimization: City-Level Multi-Datacenter Energy Saving Coordination}

\subsubsection{Problem Statement}

\textbf{Background:}

A technology company operates multiple data centers across a city. Each data center has discrete switch states (on/off) and load scheduling involves absolute value functions, leading to non-differentiable objective functions. The company needs to coordinate energy consumption across all centers to minimize global costs.

\textbf{Known Data:}
\begin{itemize}
    \item Data center locations and capacities
    \item Energy pricing structures
    \item Computing power demands
    \item Cooling requirements
    \item Switch state transition costs
\end{itemize}

\textbf{Decision Objective:}

Minimize global energy consumption costs while meeting computing power and cooling constraints for each center.

\subsubsection{Problem Formulation Validity Evaluation}

\textbf{Problem Summary:}

A multi-datacenter operator must solve a non-smooth optimization problem with discrete switch states and absolute value cost functions, requiring subgradient methods or heuristic approaches for large-scale discrete variable coordination.

\textbf{Detailed Scores:}

\begin{table}[h]
\centering
\begin{tabular}{|p{4.5cm}|c|p{6cm}|}
\hline
\textbf{Criterion} & \textbf{Score} & \textbf{Evidence} \\
\hline
TD.1 Business Semantic Authenticity & 0.9 & Uses data center operations language: ``energy saving'', ``load scheduling'', ``cooling requirements'', ``switch states''. \\
\hline
TD.2 Anti-Template Design & 0.9 & Non-differentiable objective prevents gradient-based methods; discrete states create combinatorial complexity. \\
\hline
TD.3 Multi-Constraint Coupling & 0.85 & Global cost minimization couples local decisions; cooling constraints interact with computing loads. \\
\hline
TD.4 Anti-Pattern-Matching & 0.9 & ``Energy optimization'' suggests continuous LP/QP, but problem requires non-smooth/discrete techniques. \\
\hline
\end{tabular}
\end{table}

\textbf{Overall Score: 0.89 / 1.00}

\newpage

\subsection{Sequential Decision under Contextual MNL: Cinema Scheduling Dynamic Optimization}

\subsubsection{Problem Statement}

\textbf{Background:}

A cinema chain must decide daily movie schedules. Customer choices follow a Multinomial Logit (MNL) model with unknown attraction parameters. The cinema must learn these parameters online from aggregated sales data and adjust schedules dynamically.

\textbf{Known Data:}
\begin{itemize}
    \item Available movies and showtimes
    \item Historical sales data (aggregated)
    \item MNL model structure
    \item Revenue per ticket
    \item Screening room capacities
\end{itemize}

\textbf{Decision Objective:}

Maximize cumulative expected revenue over $T$ days, balancing the conflict between ``learning parameters'' and ``instant monetization.''

\subsubsection{Problem Formulation Validity Evaluation}

\textbf{Problem Summary:}

A cinema must solve a sequential decision problem combining online parameter estimation for MNL choice models with dynamic schedule optimization, requiring joint learning and decision-making under Logit demand.

\textbf{Detailed Scores:}

\begin{table}[h]
\centering
\begin{tabular}{|p{4.5cm}|c|p{6cm}|}
\hline
\textbf{Criterion} & \textbf{Score} & \textbf{Evidence} \\
\hline
TD.1 Business Semantic Authenticity & 0.9 & Uses cinema operations language: ``movie scheduling'', ``showtimes'', ``ticket revenue'', ``screening rooms''. \\
\hline
TD.2 Anti-Template Design & 0.9 & Coupling of parameter estimation and decision-making prevents separate optimization; MNL structure requires specialized learning algorithms. \\
\hline
TD.3 Multi-Constraint Coupling & 0.8 & Room capacity constraints interact with demand learning; schedule decisions affect future information gain. \\
\hline
TD.4 Anti-Pattern-Matching & 0.9 & ``Scheduling'' suggests static optimization, but problem requires online learning with dynamic regret minimization. \\
\hline
\end{tabular}
\end{table}

\textbf{Overall Score: 0.88 / 1.00}

\subsection{Future Information Leakage: Counterfactual Reasoning in Optimization}

\subsubsection{Problem Statement}

\textbf{Background:}

This problem explores the impact on current decision boundaries if the decision-maker can foresee partial future information (e.g., demand distribution shifts). It investigates optimization under non-causal logic and quantifies the value of information advantage.

\textbf{Known Data:}
\begin{itemize}
    \item Current decision problem structure
    \item Hypothetical future information scenarios
    \item Decision maker's foresight capability
\end{itemize}

\textbf{Decision Objective:}

Quantify marginal benefits of information advantage and design regularization mechanisms to prevent overfitting to future data.

\subsubsection{Problem Formulation Validity Evaluation}

\textbf{Problem Summary:}

This frontier problem examines counterfactual reasoning in optimization, evaluating the Value of Information (VoI) when decision-makers have access to partial future knowledge, and designing safeguards against bias.

\textbf{Detailed Scores:}

\begin{table}[h]
\centering
\begin{tabular}{|p{4.5cm}|c|p{6cm}|}
\hline
\textbf{Criterion} & \textbf{Score} & \textbf{Evidence} \\
\hline
TD.1 Business Semantic Authenticity & 0.8 & Abstract problem framing; uses terms like ``information advantage'', ``decision boundaries'', ``foresight''. \\
\hline
TD.2 Anti-Template Design & 0.95 & Non-causal logic prevents standard optimization approaches; requires counterfactual reasoning frameworks. \\
\hline
TD.3 Multi-Constraint Coupling & 0.75 & Primarily conceptual; coupling arises from information structure rather than physical constraints. \\
\hline
TD.4 Anti-Pattern-Matching & 0.95 & ``Optimization'' suggests standard techniques, but problem requires philosophical/methodological innovation. \\
\hline
\end{tabular}
\end{table}

\textbf{Overall Score: 0.85 / 1.00}

\textbf{Note:} This problem represents a frontier exploration topic rather than a traditional optimization benchmark. It challenges agents to think beyond standard causal decision-making frameworks.

\end{document}